\documentclass[11pt]{article}
 
\usepackage[preprint]{acl}
 
\usepackage{times}
\usepackage{latexsym}
\usepackage[T1]{fontenc}
\usepackage[utf8]{inputenc}
\usepackage{microtype}
\usepackage{inconsolata}
\usepackage{graphicx}
\usepackage{amsmath,amssymb,amsthm,array,float,booktabs,multirow}
\usepackage{fancyvrb}
\usepackage{enumitem}
\newtheorem{proposition}{Proposition}
\floatstyle{ruled}
\newfloat{algorithm}{tbp}{loa}
\floatname{algorithm}{Algorithm}
\floatstyle{plain}
\DeclareMathOperator{\In}{In}
\DeclareMathOperator{\Out}{Out}
 
\newcommand{\ours}{CGAP}
\title{Capability-Grounded Agentic Planning}
\author{Shikha Shrestha \quad William Yeoh \quad Ning Zhang \quad Yevgeniy Vorobeychik \\
  Washington University in St.\ Louis}
 
\begin{document}
\maketitle
 
\begin{abstract}
LLM agents serve natural-language requests through tools, data sources,
and other agents, and even simple requests often need several
interdependent calls. Typically, the LLM picks these calls by trial and
error, with no guarantee of an executable plan. We present
Capability-Grounded Agentic Planning (CGAP), which instead reasons about
what a request requires and what tools provide. CGAP compiles the tool
catalog deterministically into a PDDL model, uses an LLM to turn the
request into goals, and solves the model with a classical planner. Since
tool outputs are unknown in advance, the model is abstract, tracking
which inputs are available, not their values. During execution, an LLM
adds entity-specific calls that abstract plans omit, and deterministic
checks reject any that break the plan. Every argument is drawn from a
cited source, such as a conversation span or an earlier tool response,
and each requested result counts as delivered only after an LLM verifier
checks it against the request. With three LLMs on ToolSandbox,
ComplexFuncBench, BFCL multi-turn, and $\tau^2$-bench retail, CGAP
outperforms a standard tool-calling agent by up to 28.9 points.
\end{abstract}
 
\section{Introduction}
\label{sec:intro}
 
LLM agents are increasingly used to retrieve information and act on behalf
of users through external tools \citep{yao2023react,qin2024toolllm}.
However, both the tasks to solve and the tools that solve them are
described in natural language (NL), which is imprecise and highly
flexible. The LLM that drives the agent must therefore both
\emph{understand} the request and \emph{construct an executable plan} for
it, and even simple requests require several interdependent operations.
For example, with basic forecasting, date-lookup, and geolocation tools, a
request for the ``weather in Paris and Nice next Monday, and in Nice on
Tuesday'' takes seven calls: two city lookups, two date resolutions, and
three forecasts, with the outputs of some calls supplying the inputs of
others.
 
ReAct-style agents address such problems by interleaving reasoning with
tool calls, choosing one call at a time \citep{yao2023react}, while
plan-first systems such as HuggingGPT, ReWOO, and LLMCompiler write out a
task decomposition or dependency graph before acting
\citep{shen2023hugginggpt,xu2023rewoo,kim2024llmcompiler}, a strategy that
Plan-and-Solve also applies to reasoning tasks
\citep{wang2023planandsolve}. However, both classes of approaches use the
LLM for both request interpretation and planning, so nothing guarantees
that a call's inputs will be available when it runs. LLMs also have
documented limitations on formal planning problems
\citep{valmeekam2023planbench}, and intrinsic self-correction can fail
without external feedback \citep{huang2024selfcorrect}. LLM-based agentic
workflow planning therefore remains unreliable, particularly with smaller
(and cheaper) LLMs.
 
Hybrid (neuro-symbolic) alternatives combine LLMs with formal planners.
Many, such as LLM+P, use an LLM to translate a planning problem or action
model into PDDL \citep{mcdermott1998pddl} for a classical planner
\citep{liu2023llmp,guan2023leveraging,tantakoun2025survey}, while
LLM-Modulo frameworks pair LLM proposals with external critics
\citep{kambhampati2024llmmodulo}. These approaches start from actions that
are described formally or that the LLM writes down, and a plan is only as
valid as that model. Tool-using agents instead work from \emph{NL
specifications of requests and tool capabilities}, in an \emph{open
world}: the values that their tools take and return are unknown until a
call returns, and they do not come from a finite set that a symbolic
planning model can enumerate.
 
\begin{figure*}[t]
\centering
\includegraphics[width=\textwidth]{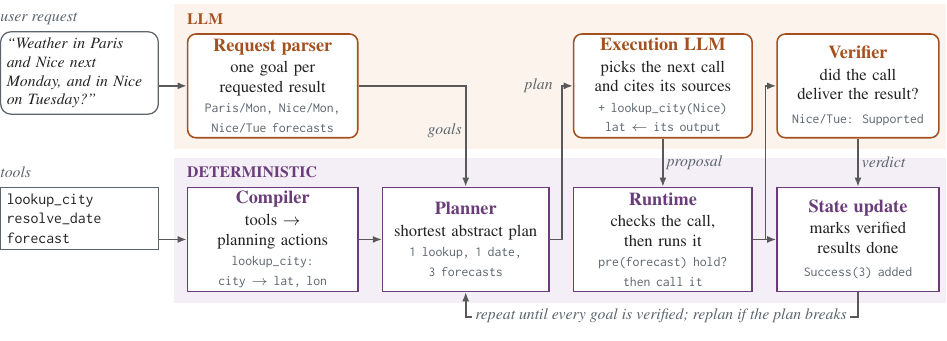}
\caption{\textbf{One user turn of \ours{}.} LLM components (top: the
request parser $M_\theta$, the execution LLM $E_\phi$, and the verifier)
interpret the request, choose each call, and judge its result.
Deterministic components (bottom) build the planning model, search for a
shortest plan, check every proposed call before it runs, and record which
results are complete. Grey text shows each component's work on the
weather request. The planner ablations (\S\ref{sec:ablations}) replace
only the planner, with an LLM or a dependency-graph planner.}
\label{fig:overview}
\end{figure*}

We propose Capability-Grounded Agentic
Planning (\ours{}), a framework for agentic workflow planning. \ours{}
reads the NL request together with the catalog of capabilities available
to the agent, such as tools, data sources, or other agents, each with an
NL description and the input schema that function-calling interfaces
require; the LLM's own abilities, such as summarizing, are not modeled as
capabilities. \ours{} compiles the catalog deterministically into a PDDL
planning model, with required inputs as preconditions and outputs as
effects (\S\ref{sec:compilation}), and an LLM parses the request into one
goal per requested result (\S\ref{sec:goals}). For open-world inputs and
outputs, the model abstracts away values: it tracks only which named
inputs and outputs (\emph{fields}) are available, a finite set fixed by
the catalog. A classical planner solves this model, and LLMs then turn the
abstract plan into executable calls and verify that their results match
the request. Our experiments test whether deriving the dependencies
between calls from this model improves tool use over having an LLM
generate them. We instantiate the planner with state-space search, and we
also evaluate a deterministic dependency-graph planner that ignores delete
effects, to see how much of the gain remains without them.
Figure~\ref{fig:overview} shows one user turn of this process.
 
In the weather example, the planner cannot predict what the tools will
return, so the planning state records only whether
a value \emph{is available} for a field such as \texttt{latitude}, an
output of the city lookup and a required input of the forecast, and not
which city it describes. Each requested forecast is a separate goal, but
one city lookup makes coordinates available for every forecast, so the
planner returns a five-action abstract plan for a request that takes
seven calls (Figure~\ref{fig:pipeline}). An execution LLM proposes the
entity-specific calls the plan omits, here a lookup for Nice and a second
date resolution, and a deterministic \emph{runtime}, which checks every
call before it runs, accepts such a proposal only if the remaining plan
stays valid. Every argument must cite its source, so each coordinate can
be traced to the lookup that produced it, and a forecast counts as
delivered only after deterministic checks and an LLM verifier accept the
call that produced it.
 
We make three contributions:
\begin{itemize}[leftmargin=*,nosep]
\item \textbf{Neuro-symbolic agentic planning.} \ours{}, a hybrid approach
that compiles tool catalogs deterministically into an abstract,
field-level PDDL model that can be solved before tool outputs are known,
uses an LLM to turn the request into its goals, and solves the model with
classical planning (\S\ref{sec:compilation}, \S\ref{sec:goals}).
\item \textbf{Checked execution with provable guarantees.} A runtime that
validates the entity-specific calls LLMs add, instantiates arguments from
cited sources, and verifies each requested result; every execution it
admits maps onto a valid abstract plan, so exhaustive search failure
certifies infeasibility for the compiled catalog (\S\ref{sec:guarantees}).
\item \textbf{Experiments on four tool-using agentic benchmarks.} With
three LLMs on ToolSandbox, ComplexFuncBench, BFCL multi-turn, and
$\tau^2$-bench retail, \ours{} outperforms a standard tool-calling agent in
all 15 model--metric comparisons and, with gpt-oss-20b, four further
baselines. Replacing deterministic dependency planning with LLM-generated
plans costs up to 16.8 points, less with GPT-5 mini; a deterministic graph
planner retains most of the gain, and search adds to it where actions have
delete effects (\S\ref{sec:experiments}).
\end{itemize}
 
\paragraph{Related work.}
A common way to organize tool use is a graph of tool dependencies. TaskBench
and ToolNet build such graphs explicitly \citep{shen2024taskbench,liu2024toolnet},
\citet{wu2024graph} plan on them with graph neural networks, and
NaviAgent and GRAFT add execution feedback and learned dependencies,
respectively \citep{jiang2026naviagent,graft2026}.
\ours{} instead derives dependencies deterministically from capability
interfaces and lets them govern execution through validated
entity-specific refinements, source-grounded arguments, and
per-operation completion records. Our graph-planning variant shares this
model and runtime, so comparing it with search isolates the value of
reasoning about delete effects, such as a call that switches off a
setting a later call needs (\S\ref{sec:ablations}). The planners
closest to ours are the lifted regression methods LLM-Regress and NL-PDDL,
which use an LLM to supply commonsense knowledge during planning
\citep{liu2025llmregress,liu2026nlpddl}; they keep entities as variables,
whereas \ours{} drops entity distinctions and leaves them to execution.
Agent+P likewise pairs a UI transition graph with a symbolic planner
\citep{ma2025agentp}.
Appendix~\ref{app:related} relates \ours{} to service composition, API
synthesis, executable request representations, attribution, and
prompt-injection defenses.
 
\section{Capability-Grounded Planning}
\label{sec:method}
 
Figure~\ref{fig:overview} shows one user turn of \ours{}; the weather
request serves as a running example throughout.
 
\subsection{Setting}
\label{sec:problem}
 
A user interacts with an agent over $T$ turns. Each turn $t$ opens with an
instruction $q_t$, which may request several results and may extend,
revise, or cancel earlier requests; the agent then makes calls and
replies. A \emph{call} is one invocation of a capability together with
its response. We write $H_t$ for the history before turn $t$,
$H_t^{\mathrm{user}}$ for the user messages in it, and $H_{t,k}$ for $H_t$
followed by $q_t$ and the first $k$ calls of turn $t$. The agent cannot
observe whether the user finds the outcome acceptable, so it tracks a
condition it can check: every requested operation has a call that was
verified against the request.
 
The catalog $\mathcal C$ lists \emph{capabilities}, such as tools, data
sources, API operations, or other agents. Each has an NL description and an input
schema, which, as in function-calling interfaces, names its parameters
(e.g., \texttt{city}); some also have a return schema, documented
connections to other tools, or preconditions and effects. The
\emph{application}, here a benchmark adapter, supplies the catalog,
annotations, observation rules, transformations, and authorization
rules. For a capability $c$, we distinguish its \emph{action} $a_c$, a
symbolic counterpart without arguments, from an
\emph{invocation} $c(\mathbf v)$, which calls the same $c$ with concrete
arguments and returns a \emph{response}.
 
\subsection{Compiling the catalog into a planning model}
\label{sec:representation}
\label{sec:compilation}
 
To order calls, the planner must know which capabilities produce the
inputs of others; \ours{} reads this off the parameter names in the
catalog's schemas, so the catalog needs no planning syntax.
 
\paragraph{Facts.}
Let $\mathcal X$ be the set of \emph{fields}, the names of the inputs and
outputs that the catalog's schemas declare (e.g., \texttt{city},
\texttt{lat}, \texttt{date}), after a fixed normalization
(Appendix~\ref{app:action-construction}). Function-calling catalogs
always name their inputs; outputs are named by return schemas where given
and are otherwise inferred, as described next. The fact
$\operatorname{Available}(p)$ holds when some value for field $p$ has a
\emph{source}, that is, a span of a user message or an expected output
field of a successful response (R1, \S\ref{sec:execution}). Application-supplied domain predicates describe other
conditions, such as the observed state of a device setting, and
$\operatorname{Success}(i)$ records that requested operation $i$ is
complete. We write $\mathcal F$ for the set of facts; a state
$s\subseteq\mathcal F$ is the set of facts that hold.
 
\paragraph{Actions.}
For a capability $c$, let $\In(c)\subseteq\mathcal X$ be the required
input fields of its schema and $\Out(c)\subseteq\mathcal X$ its candidate
outputs. Candidate outputs come from the return schema when one is given,
and otherwise from naming conventions and documented tool connections
(Appendix~\ref{app:action-construction}). Optional annotations
$K_c^{\mathrm{pre}},K_c^+,K_c^-$ supply domain preconditions, add effects,
and delete effects. Writing
$\operatorname{Av}(P)=\{\operatorname{Available}(p)\mid p\in P\}$ for
$P\subseteq\mathcal X$, each capability compiles deterministically to a \emph{base action} $a_c$:
\begin{align}
 \operatorname{pre}(a_c)&=\operatorname{Av}(\In(c))\cup K_c^{\mathrm{pre}},\nonumber\\
 \operatorname{add}(a_c)&=\operatorname{Av}(\Out(c))\cup K_c^+,\label{eq:compile}\\
 \operatorname{del}(a_c)&=K_c^-,\nonumber
\end{align}
and $\mathcal A_{\mathcal C}=\{a_c\mid c\in\mathcal C\}$. Candidate
outputs describe only what $c$ \emph{may} return; at run time, a field
becomes available only when a response actually contains it. An action
\emph{achieves} $f$ if $f\in\operatorname{add}(a)$, and
$\operatorname{Ach}(f)$ is the set of base actions that achieve $f$;
predicates must match exactly after normalization. In the running example,
\texttt{lookup\_city} achieves the coordinate inputs of \texttt{forecast},
and \texttt{resolve\_date} achieves its \texttt{date}.
 
Domain predicates capture prerequisites and state changes that input and
output schemas do not express. \ours{} takes these rules as
application-supplied modeling inputs; automatically inducing them from
natural-language descriptions is outside our scope. The annotations
describe reusable capability behavior, leaving the planner to determine
the action sequence for each request. In ToolSandbox, for example,
sending a message requires cellular service \citep{lu2025toolsandbox}.
The messaging action therefore requires an observation of the cellular
state, whose value the runtime checks before invocation. Status setters
compile to one action per value, with delete effects removing the
opposite status fact (Appendix~\ref{app:action-construction}).
 
\paragraph{Plans.}
With the PDDL successor function
$\gamma(s,a)=(s\setminus\operatorname{del}(a))\cup\operatorname{add}(a)$
\citep{mcdermott1998pddl,ghallab2004automated}, a plan
$\langle a_1,\ldots,a_n\rangle$ is valid for
$\mathcal P=\langle\mathcal F,\mathcal A,I,G\rangle$ if $s_0=I$,
$G\subseteq s_n$, and for $1\leq\ell\leq n$,
\begin{equation}
 \operatorname{pre}(a_\ell)\subseteq s_{\ell-1},\qquad
 s_\ell=\gamma(s_{\ell-1},a_\ell).
 \label{eq:valid-plan}
\end{equation}
 
\paragraph{Abstract plans.}
The state records that a latitude is available, not whose latitude it is:
coordinates for Paris and for Nice both establish
$\operatorname{Available}(\texttt{lat})$, and the values and their sources
stay in the conversation history. This is deliberate: values are unknown
before a call, so a value-indexed state would force the planner to predict
responses, and catalogs say what a capability returns, not which request a
returned value will serve. It also keeps the fact vocabulary independent
of the number of entities, as in generalized planning, which seeks plans
that hold for any number of objects \citep{silver2024generalized},
although \ours{} still plans for each request. Following the principle of
planning abstractions \citep{sacerdoti1974abstrips,knoblock1991characterizing},
we call a plan over this model an \emph{abstract plan}; its actions are
capabilities without arguments, not composite tasks as in HTN planning. It
fixes which requested results must be
produced, which kinds of prerequisites each needs, and in what order, but
not how many copies of a supporting call distinct entities require. This
is why the weather plan has three forecasts but one lookup: each requested
result is a goal of its own (\S\ref{sec:goals}), whereas one lookup makes
\texttt{lat} available for every city. Execution supplies the missing
copies under validation (\S\ref{sec:execution}), and
Section~\ref{sec:guarantees} shows that every compliant execution still
maps onto a valid abstract plan.
 
\subsection{Parsing requests into operations}
\label{sec:goals}
 
\begin{figure*}[t]
\centering
\includegraphics[width=\textwidth]{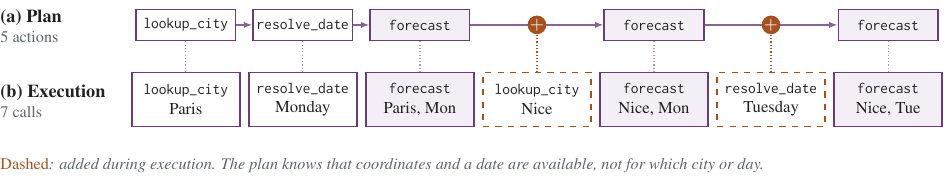}
\caption{\textbf{Why the abstract plan is shorter than its execution.}
(a)~Search plans one city lookup and one date resolution, which make every
input of the three requested forecasts (shaded) available. (b)~During
execution, $E_\phi$ adds a lookup for Nice and a date resolution for
Tuesday (dashed) through validated refinements, so five planned actions
become seven calls.}
\label{fig:pipeline}
\end{figure*}
 
\paragraph{Request parsing.}
At the start of turn $t$, an LLM $M_\theta$ receives the instruction
$q_t$, the user history $H_t^{\mathrm{user}}$, the previous operations
$\mathcal O_{t-1}$, domain instructions $d$, and the names, descriptions,
and input schemas $\mathcal C_{\mathrm{in}}$ of the capabilities. It
returns the requested operations $\mathcal O_t$, user-supplied input
values $\mathcal V_t$, global constraints $\Lambda_t$, and requests $U_t$
that no capability can serve:
\begin{equation}
 (\mathcal O_t,\mathcal V_t,\Lambda_t,U_t)
 =M_\theta(q_t,H_t^{\mathrm{user}},\mathcal O_{t-1},d,\mathcal C_{\mathrm{in}}).
 \label{eq:mapping}
\end{equation}
An operation
$o_i=(i,c_i,r_i,\epsilon_i,\zeta_i)$ names a capability $c_i$, a
requirement $r_i$ (entity, time, and scope), evidence spans $\epsilon_i$
in the user's messages, and a status
$\zeta_i\in\{\mathrm{active},\mathrm{cancelled}\}$. What $M_\theta$ must
produce is deliberately small: one operation per requested result,
including those still unfinished from earlier turns, and no prerequisite
steps. In the
running example, $M_\theta$ returns three forecast operations and leaves
the calendar dates to \texttt{resolve\_date}. A validator checks span
bounds and literal, schema-compatible value extraction, and allows at most
one repair. Whether a span actually expresses the intended operation
remains a judgment of the LLM (Appendix~\ref{app:prompt}).
 
\paragraph{Terminal actions and the planning problem.}
Each requested result needs its own goal, so for each active $o_i$ we add
a \emph{terminal action} $a_i^{\mathrm{term}}$ with the preconditions and
delete effects of $a_{c_i}$ and
$\operatorname{add}(a_i^{\mathrm{term}})=\operatorname{add}(a_{c_i})\cup\{\operatorname{Success}(i)\}$.
The terminal action is the only achiever of
$\operatorname{Success}(i)$, so a result already in the history, such as a
forecast from an earlier turn, cannot complete a newly requested
operation without a new call, and three requested forecasts yield three separate goals. Data in
the history can still supply \emph{inputs}: coordinates looked up for one
forecast serve every later call that needs them. The LLM thus specifies the requested results, and search
determines their dependencies at the field level. After $k$ calls of turn
$t$, the planning problem is
$\mathcal P_{t,k}=\langle\mathcal F_t,\mathcal A_t,I_{t,k},G_t\rangle$,
with
$\mathcal A_t=\mathcal A_{\mathcal C}\cup\{a_i^{\mathrm{term}}\}_{o_i\,\mathrm{active}}$,
$G_t=\{\operatorname{Success}(i)\}_{o_i\,\mathrm{active}}$, and
\begin{equation}
 I_{t,k}=\operatorname{Av}(X_{t,k})\cup W_{t,k}\cup D_{t,k},
 \label{eq:problem}
\end{equation}
where $X_{t,k}$ is the set of available fields, $W_{t,k}$ the domain
facts maintained by the application's observation rules, and $D_{t,k}$ the
completion markers backed by accepted completion records;
Equation~\ref{eq:observation-update} gives how each call updates these
sets. Values never enter the state. In the running example,
$I_{0,0}=\operatorname{Av}(\{\texttt{city},\texttt{day}\})$ and
$G_0=\{\operatorname{Success}(1),\operatorname{Success}(2),\operatorname{Success}(3)\}$.
 
\subsection{Planning and grounded execution}
\label{sec:planning}
\label{sec:execution}
 
Within a turn, search and execution alternate
(Algorithm~\ref{alg:framework}) under three budgets: $B$ state expansions
per search, $L$ actions per refinement, and $N$ execution decisions per
turn. These are hyperparameters, fixed in advance and shared by all
benchmarks and LLMs ($B=2{,}048$, $L=4$, $N=32$).
 
\paragraph{Search and execution decisions.}
Breadth-first search solves $\mathcal P_{t,k}$
(Algorithm~\ref{alg:classical}) and returns a shortest abstract plan,
\textsc{NoPlan} if it has expanded every reachable state without reaching
$G_t$, or \textsc{Limit} if it has expanded a predefined number $B$ of
states first. In the running example, the plan has one city lookup, one
date resolution, and three forecasts (Figure~\ref{fig:pipeline}(a)).
Execution then works through this plan; after $k$ calls, the
\emph{remaining plan} $\bar\pi_{t,k}$ is the part of it not yet executed.
Given $\bar\pi_{t,k}$, an execution LLM $E_\phi$ proposes one of three
decisions, each counting toward $N$: 1)~\textsc{Execute}, which supplies
argument sources for the first planned action; 2)~\textsc{Refine}, which
proposes supporting actions $\rho$ to prepend when the plan needs an input
for which no source exists; and 3)~\textsc{Stop}, which asks the user for
clarification or reports work that cannot be done. We describe
\textsc{Execute} and \textsc{Refine} next.
 
The LLM never writes an argument value. Each argument of an
\textsc{Execute} decision cites a source, a span of the conversation or a
field of a successful response, and a \emph{resolver} from a fixed set
that the LLM cannot extend: a \emph{decoder} copies the cited text or
converts it to the schema's type (e.g., \texttt{"2"} to the integer 2),
and a \emph{transformation} is a deterministic function supplied by the
application, such as a date converter. Application defaults are also
allowed. Before calling, the
runtime checks the action's preconditions in $I_{t,k}$, the argument
schema, the sources, and the application's authorization rules. Whether a
sourced argument refers to the intended entity is assessed after the call
by the completion verifier, so for tools with side effects this check
cannot undo a wrongly targeted action.
 
Refinement (\textsc{Refine}) is how an abstract plan becomes a concrete
one. After the Paris
forecast, $\operatorname{Available}(\texttt{lat})$ still holds, but no
source contains the coordinates of Nice. $E_\phi$ therefore prepends a
lookup for Nice, and later a date resolution for Tuesday, so that five
planned actions become seven calls (Figure~\ref{fig:pipeline}(b)). A
refinement holds at most $L$ actions, each a recursive achiever of a
precondition of a pending terminal action, and no terminal actions. It is
kept only if $\rho\mathbin{\|}\bar\pi_{t,k}$, the prepended actions
followed by the remaining plan, is valid from $I_{t,k}$
(Appendix~\ref{app:execution-validation}). A refinement can therefore add
supporting calls, but it cannot add, remove, or reorder the terminal
actions of the current plan; only search can reorder them.
Validation also cannot tell whether a refinement is \emph{needed}: a
redundant second lookup for Paris passes as easily as the lookup for Nice,
so necessity is left to the LLM. Without delete effects
(ComplexFuncBench, BFCL, $\tau^2$-bench retail), added facts cannot break
the remaining plan, so only the prefix is checked.
 
\paragraph{Completion verification.}
A call \emph{succeeds} if the tool reports success; whether it delivers
its operation's result is decided separately. For each successful terminal
call, deterministic checks confirm its capability, sources, and
application-supplied domain tests, and an LLM verifier compares the execution record $e$ with
$r_i$, $\epsilon_i$, and $\Lambda_t$. The verifier returns
\textsc{Supported}, \textsc{Contradicted}, or \textsc{Unknown} with cited
evidence. Only an accepted record, one with passing checks and a
\textsc{Supported} verdict, adds $\operatorname{Success}(i)$, and a call
completes at most one operation. The verifier should reject, e.g., a
forecast for the wrong city, but it can err.
 
\paragraph{State update.}
A call $e$ adds facts $O_e^+$, namely newly available fields, established
domain facts, and $\operatorname{Success}(i)$ if its completion record for
$o_i$ is accepted, and deletes facts $O_e^-$:
\begin{equation}
 I_{t,k+1}=(I_{t,k}\setminus O_e^-)\cup O_e^+.
 \label{eq:observation-update}
\end{equation}
Equation~\ref{eq:problem} gives the parts of the state, and
Equation~\ref{eq:observation-update} how a call changes them: the parts
use disjoint predicates, so it updates $X$, $W$, and $D$ separately, and
$I_{t,k+1}$ keeps the form of Equation~\ref{eq:problem}. The runtime
enforces three rules (Appendix~\ref{app:observation}). \emph{(R1)} A successful response of $c$
makes available only fields in $\Out(c)$; other parsed fields can be cited
as values but make nothing available. \emph{(R2)} A successful call of $c$
deletes all of $K_c^-$ and adds no domain fact outside $K_c^+$.
\emph{(R3)} A failed call may delete facts but adds none. Search reruns
if the remaining plan becomes invalid or a call fails, lacks a
required output, or has its completion rejected. A turn ends when
$G_t\subseteq I_{t,k}$, at a \textsc{Stop} decision, when search returns
\textsc{NoPlan} or \textsc{Limit}, or after $N$ decisions; $E_\phi$ then
replies from the accepted records and the unresolved work
(Algorithm~\ref{alg:framework}).
 
\subsection{What the planner and the runtime certify}
\label{sec:guarantees}
 
Proposition~\ref{prop:accounting} follows from the runtime checks alone,
so it also holds for the planner ablations (\S\ref{sec:ablations}), as
does Proposition~\ref{prop:plan}(b); only exhaustive search, however,
certifies that no plan exists.
 
\begin{proposition}[Runtime invariants]
\label{prop:accounting}
\label{prop:admission}
In every turn $t$ of Algorithm~\ref{alg:framework} and for every $k$:
\begin{enumerate}[leftmargin=2.2em,nosep]
\item[(i)] \emph{Applicability.} Every called action's preconditions hold
in $I_{t,k}$ at the time of the call, and the terminal actions of the remaining plan
are, in order, a suffix of those in the plan last returned by search.
\item[(ii)] \emph{Provenance.} Every argument value is computed by a
decoder or an application-supplied transformation from user spans or
successful-response fields in $H_{t,k}$, or is a default the application
configured for that argument.
\item[(iii)] \emph{Completion accounting.} Every
$\operatorname{Success}(i)\in D_{t,k}$ is backed by an accepted completion
record for a successful call of $c_i$, made in turn $t$ or inherited from
an unchanged operation, and distinct markers by distinct calls.
\end{enumerate}
\end{proposition}
 
These structural invariants are proved in Appendix~\ref{app:proof}. A
cited source, transformation, or default may still yield a wrong value,
the verifier that accepted a record may be wrong, and the free-form final
reply is not constrained. 
 
\begin{table*}[t]
\centering
\small
\setlength{\tabcolsep}{4.5pt}
\begin{tabular}{@{}llccccc@{}}
\toprule
 & & & \multicolumn{2}{c}{ComplexFuncBench} & & \\
\cmidrule(lr){4-5}
Model & Method & ToolSandbox & Success & Call Acc. & BFCL-MT & $\tau^2$-Retail \\
\midrule
\multirow{6}{*}{gpt-oss-20b} & Plain agent & 52.3\,{\scriptsize[47.2, 57.5]} & 47.6\,{\scriptsize[41.6, 53.6]} & 66.0\,{\scriptsize[61.6, 70.5]} & 34.0\,{\scriptsize[27.5, 40.5]} & 47.4\,{\scriptsize[38.6, 56.1]} \\
 & \quad + dependency notes & 58.4\,{\scriptsize[53.2, 63.3]} & -- & -- & -- & -- \\
 & \quad compute-matched & 57.1\,{\scriptsize[52.0, 62.2]} & 50.8\,{\scriptsize[45.2, 56.4]} & -- & 37.0\,{\scriptsize[30.5, 43.5]} & 51.8\textsuperscript{\dag}\,{\scriptsize[43.0, 60.5]} \\
 & ReWOO & 54.1\,{\scriptsize[47.5, 60.6]} & 46.4\,{\scriptsize[41.2, 51.6]} & -- & 33.5\,{\scriptsize[27.0, 40.0]} & 44.7\,{\scriptsize[36.0, 53.5]} \\
 & LLMCompiler & 54.2\,{\scriptsize[47.7, 60.6]} & 40.4\,{\scriptsize[34.4, 46.4]} & 50.7\,{\scriptsize[46.2, 55.2]} & 21.0\,{\scriptsize[15.5, 27.0]} & 30.7\,{\scriptsize[22.8, 39.5]} \\
 & \ours{} (ours) & \textbf{81.2}\,{\scriptsize[77.4, 84.8]} & \textbf{58.0}\,{\scriptsize[52.0, 64.0]} & \textbf{74.9}\,{\scriptsize[70.8, 79.1]} & \textbf{52.0}\,{\scriptsize[45.0, 59.0]} & \textbf{61.4}\,{\scriptsize[52.6, 70.2]} \\
\midrule
\multirow{2}{*}{Qwen3-32B} & Plain agent & 58.6\,{\scriptsize[53.5, 63.7]} & 53.2\,{\scriptsize[47.2, 59.2]} & 72.2\,{\scriptsize[67.9, 76.4]} & 48.0\,{\scriptsize[41.0, 55.0]} & 33.3\,{\scriptsize[24.6, 42.1]} \\
 & \ours{} (ours) & \textbf{83.1}\,{\scriptsize[79.5, 86.5]} & \textbf{64.0}\,{\scriptsize[58.0, 70.0]} & \textbf{81.3}\,{\scriptsize[77.4, 85.2]} & \textbf{62.0}\,{\scriptsize[55.0, 68.5]} & \textbf{44.7}\,{\scriptsize[36.0, 53.5]} \\
\midrule
\multirow{2}{*}{GPT-5 mini} & Plain agent & 72.4\,{\scriptsize[68.0, 76.7]} & 60.8\,{\scriptsize[54.8, 66.8]} & 81.9\,{\scriptsize[78.3, 85.5]} & 56.5\,{\scriptsize[49.5, 63.5]} & 64.9\,{\scriptsize[56.1, 73.7]} \\
 & \ours{} (ours) & \textbf{86.3}\,{\scriptsize[83.1, 89.2]} & \textbf{66.0}\,{\scriptsize[60.0, 72.0]} & \textbf{87.0}\,{\scriptsize[83.6, 90.3]} & \textbf{62.5}\,{\scriptsize[55.5, 69.0]} & \textbf{73.7}\,{\scriptsize[65.8, 81.6]} \\
\bottomrule
\end{tabular}
\caption{\textbf{Main results} (\%). ToolSandbox reports the average
milestone similarity score, ComplexFuncBench the success rate and call
accuracy, BFCL-MT the accuracy on BFCL multi-turn, and $\tau^2$-Retail
the success rate on all 114 $\tau^2$-bench retail tasks, over three runs
per task (\S\ref{sec:setup}). Brackets give 95\% bootstrap intervals over
tasks. The four additional baselines run with gpt-oss-20b only, the
dependency-notes agent on ToolSandbox only, and three lack call accuracy
(--); \dag\,without a token budget (Appendix~\ref{app:extra-baselines}).
Bold: best per model; Table~\ref{tab:gains} gives paired differences.}
\label{tab:main}
\end{table*}
 
\begin{proposition}[Executions refine abstract plans]
\label{prop:plan}
\label{prop:search}
\emph{(a)} Every plan that Algorithm~\ref{alg:classical} returns on
$\mathcal P_{t,k}$ is valid and has the fewest actions among valid plans.
\emph{(b)} Map each successful call of $c$ to $a_i^{\mathrm{term}}$ if it
yields an accepted completion record for $o_i$, and to $a_c$ otherwise.
Under R1--R3, the image of every execution of
Algorithm~\ref{alg:framework} from $I_{t,k}$ that reaches $G_t$, whatever
the LLMs propose, is a valid plan for $\mathcal P_{t,k}$, so the execution
makes at least $|\bar\pi_{t,k}|$ successful calls if search returns
$\bar\pi_{t,k}$.
\emph{(c)} If search returns \textup{\textsc{NoPlan}}, for any budget $B$,
no execution of Algorithm~\ref{alg:framework} from $I_{t,k}$ completes
the turn.
\end{proposition}
 
Part (a) is standard \citep{russell2003aima}, and part (b) is the
upward-solution property of abstraction hierarchies
\citep{tenenberg1988abstraction,knoblock1991characterizing}
(Appendix~\ref{app:proof}); what is new is that the runtime enforces
R1--R3 against arbitrary LLM proposals. Through (c), \textsc{NoPlan},
unlike \textsc{Limit}, certifies that no call sequence passing the
runtime's checks can complete the request, but only for the compiled
model: an undeclared output cannot enable an action (R1). The count in (b) is loose on multi-entity requests, and we do not
use it.

\section{Experiments}
\label{sec:experiments}

\subsection{Experimental setup}
\label{sec:setup}
 
\paragraph{Benchmarks.}
We use four interactive benchmarks
(Appendix~\ref{app:benchmark-contracts}). The first, ToolSandbox
\citep{lu2025toolsandbox}, contains stateful, conversational scenarios
scored by their similarity to annotated milestones; in its
\emph{state-dependency} category, a tool works only in some world states
(sending a message requires cellular service to be on). The second,
ComplexFuncBench \citep{zhong2025complexfuncbench}, contains multi-step
requests with user constraints over real travel APIs in five domains,
scored by success rate and call accuracy. The third, BFCL multi-turn
\citep{patil2025bfcl}, checks the final environment state and the
executed calls in multi-turn conversations (base, missing functions,
missing parameters, long context). From these three
we use stratified subsets of 120 ToolSandbox scenarios, 250
ComplexFuncBench requests, and 200 BFCL conversations (50 per category).
The fourth is the retail domain of $\tau^2$-bench \citep{barres2026tau2},
a revision of $\tau$-bench's \citep{yao2025taubench}, whose 114 tasks we
all run: a customer simulated by GPT-4.1
\citep{openai2025gpt41} asks the agent to cancel, modify, return, or
exchange orders under a written policy, and a task succeeds if the final
database matches the reference one. As on ComplexFuncBench and BFCL, we
add no domain annotations: the policy's rules, such as cancelling only
pending orders, are enforced by the tools, not the planning model.
 
\paragraph{Baselines.}
We compare \ours{} with a \emph{plain agent}, which receives the tool
catalog and calls tools in a loop until it replies, under three LLMs:
gpt-oss-20b \citep{openai2025gptoss}, Qwen3-32B \citep{yang2025qwen3}, and
GPT-5 mini \citep{openai2025gpt5}. With gpt-oss-20b, four further
baselines test other explanations for \ours{}'s gains
(Appendix~\ref{app:extra-baselines}): the plain agent given our
ToolSandbox dependency annotations as prompt notes, a
\emph{compute-matched} plain agent with self-checks, and adapted versions
of two established plan-first agents, ReWOO \citep{xu2023rewoo} and
LLMCompiler \citep{kim2024llmcompiler}.
 
\paragraph{Evaluation.}
We run each configuration three times per task. Binary success metrics are
aggregated by majority (a task counts as solved if at least two runs solve
it), and ToolSandbox scores and call accuracy are averaged over runs.
Confidence intervals use 10,000 task-level bootstrap resamples, paired by
task for differences between methods, and we interpret them as pointwise
intervals; Appendices~\ref{app:benchmark-contracts}
and~\ref{app:cost-protocol} give details.
 
\begin{table*}[t]
\centering
\small
\setlength{\tabcolsep}{4.5pt}
\begin{tabular}{@{}lccccc@{}}
\toprule
 & \multicolumn{2}{c}{ToolSandbox} & & & \\
\cmidrule(lr){2-3}
Variant & Avg. & State & CFB & BFCL-MT & $\tau^2$-Retail \\
\midrule
\ours{} (search) & 81.2 & 98.4 & 58.0 & 52.0 & 61.4 \\
\midrule
LLM planning & $-$16.8\,{\scriptsize[$-$20.5, $-$13.2]} & $-$27.7\,{\scriptsize[$-$33.0, $-$22.4]} & $-$4.4\,{\scriptsize[$-$8.0, $-$0.8]} & $-$10.0\,{\scriptsize[$-$16.0, $-$4.0]} & $-$7.9\,{\scriptsize[$-$14.9, $-$0.9]} \\
graph planning & $-$3.1\,{\scriptsize[$-$6.0, +0.4]} & $-$8.9\,{\scriptsize[$-$14.5, $-$3.6]} & $-$0.4\,{\scriptsize[$-$2.8, +2.0]} & +0.5\,{\scriptsize[$-$2.5, +3.5]} & $-$1.8\,{\scriptsize[$-$7.0, +3.5]} \\
$-$ provenance & $-$5.4\,{\scriptsize[$-$8.3, $-$2.4]} & $-$4.3\,{\scriptsize[$-$8.1, $-$0.8]} & $-$3.6\,{\scriptsize[$-$7.2, $-$0.4]} & $-$5.0\,{\scriptsize[$-$9.5, $-$1.0]} & $-$5.3\,{\scriptsize[$-$11.4, +0.9]} \\
$-$ verifier & $-$2.7\,{\scriptsize[$-$4.9, $-$0.6]} & $-$1.8\,{\scriptsize[$-$4.4, +0.6]} & $-$1.6\,{\scriptsize[$-$4.0, +0.4]} & $-$2.0\,{\scriptsize[$-$5.0, +0.5]} & $-$2.6\,{\scriptsize[$-$7.9, +2.6]} \\
\bottomrule
\end{tabular}
\caption{\textbf{Ablations with gpt-oss-20b}: changes from \ours{} with
search (95\% paired bootstrap intervals). The first two rows replace the
planner; the last two remove a runtime component.}
\label{tab:ablation}
\end{table*}
 
\subsection{Main results}
\label{sec:main-results}
 
In Table~\ref{tab:main}, \ours{} outperforms every baseline on every
benchmark and metric with all three LLMs, and all 15 paired intervals
against the plain agent exclude zero (Table~\ref{tab:gains}). The gains
are smallest with GPT-5 mini (5.1--13.9 points), whose plain agent is the
strongest on every metric.
 
On ToolSandbox, \ours{} gains 13.9--28.9 points, most on state-dependency
scenarios (98.4 against 62.8 with gpt-oss-20b;
Table~\ref{tab:categories-full}), which use our hand-written
device-dependency annotations. On ComplexFuncBench, it improves success by 5.2--10.8
points and call accuracy by 5.1--9.1, and on BFCL multi-turn it gains
6.0--18.0 points. On $\tau^2$-bench retail, it gains 8.8--14.0 points although
every agent receives the same policy, both on the 64 tasks involving two
or more orders and on the other 50 (Table~\ref{tab:retail-split}).
 
\paragraph{Additional baselines.}
Dependency notes close about half of the state-dependency gap
(\S\ref{sec:ablations}). With self-checks and tokens per task within 5\%
of \ours{}'s (17\% more on $\tau^2$-bench retail), the
plain agent gains only 3.0--6.1 points; adapted ReWOO stays within 2.7
points of it, and adapted LLMCompiler is 7.2--16.7 points below it
outside ToolSandbox. \ours{} leads all four on every measure, with
paired intervals excluding zero (Table~\ref{tab:extra-paired}).
 
\subsection{Ablations}
\label{sec:ablations}
 
\subsubsection{The value of symbolic planning}
 
\begin{figure}[t]
\centering
\includegraphics[width=\columnwidth]{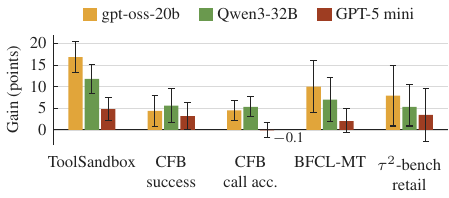}
\caption{\textbf{Deterministic planning versus LLM planning.} Gain of
\ours{} (with search) over its LLM-planning ablation for each LLM and
metric, with 95\% paired bootstrap intervals (Table~\ref{tab:gains}).}
\label{fig:gains}
\end{figure}
 
\paragraph{LLM-planning ablation.}
Our main ablation replaces every search call with one LLM call that
receives the same planning problem and returns an ordered list of actions
without arguments. The runtime validates it (Equation~\ref{eq:valid-plan})
and allows one repair, after which an invalid plan ends the turn as
\textsc{InvalidPlan}. All else is shared (Appendix~\ref{app:cost-protocol}),
so the comparison measures the effect of replacing symbolic with LLM-based
planning within the \ours{} architecture. With gpt-oss-20b and Qwen3-32B,
search improves every metric (Figure~\ref{fig:gains}), by 4.4--16.8 points, and all ten paired
intervals exclude zero. With GPT-5 mini, the gains shrink to $-$0.1 to 4.8
points, and only the ToolSandbox interval excludes zero
(Table~\ref{tab:gains}). LLM planning is still a strong variant, beating
the compute-matched agent and both plan-first agents
(Table~\ref{tab:extra-ci}); Appendix~\ref{app:llm-planning-failures}
analyzes its failures.
 
\paragraph{Graph-planning ablation.}
A second ablation replaces search with a deterministic dependency-graph
planner in the style of
TaskBench and ToolNet, which backward-chains from the pending terminal
actions over the same compiled model and ignores delete effects
(Appendix~\ref{app:cost-protocol}). With gpt-oss-20b, it scores above LLM
planning on every measure (Table~\ref{tab:ablation}; e.g., 78.1 against
64.4 on ToolSandbox), though we lack paired intervals for that
comparison. Against search, its paired intervals include zero on all
four benchmarks, and search clearly leads only on state-dependency
scenarios. On that category, access to the dependency rules does not
explain \ours{}'s score by itself: given the same rules as prompt notes,
the plain agent's score rises from 62.8 to 79.6, still 18.8 points
$[12.0, 25.7]$ below \ours{}'s 98.4. With the same compiled model, search
also beats the graph planner there by 8.9 points $[3.6, 14.5]$, which is
what modeling delete effects adds. On benchmarks without delete effects,
deterministic dependency planning accounts for most of the benefit, and
search adds no clear gain.

\subsubsection{Provenance and verification}
 
Removing provenance (Table~\ref{tab:ablation}) lowers the four
benchmark-level scores by 3.6--5.4 points (intervals exclude zero except
on $\tau^2$-bench retail), and removing verification by 1.6--2.7 (only the
ToolSandbox average excludes zero). Both mainly serve
Proposition~\ref{prop:accounting}, which success metrics do not measure.
 
\section{Conclusion}
\label{sec:conclusion}
 
We presented \ours{}, a neuro-symbolic approach in which LLMs interpret
requests, ground arguments, and judge results, while a classical planner
orders calls over an abstract model compiled from the tool catalog. On
four benchmarks, it outperforms plain tool calling with all three LLMs;
LLM planning in place of its deterministic planner lowers 14 of 15
model--metric scores (clearly with the open models, slightly with GPT-5
mini), a dependency-graph planner retains most of the benefit, and search
adds to it where actions have delete effects.
 
\section*{Limitations}
 
Our guarantees hold only for the compiled model and runtime rules:
\textsc{NoPlan} certifies infeasibility only for the catalog as compiled,
and because the abstraction merges entities, the call count of
Proposition~\ref{prop:plan}(b) is loose on multi-entity requests. How
often turns end in \textsc{NoPlan} rather than \textsc{Limit}, and hence
the certificate's practical value, is unmeasured. Request parsing can
misread an operation, a cited source can be wrong or tied to the wrong
entity, and the verifier can accept an incorrect result and, running
after the call, cannot undo a wrong side effect; we have not audited how
often these errors, or unnecessary refinements, occur. Name-based field
matching can miss or conflate connections, and breadth-first search is
exponential in the number of facts, so its budget can stop it before a
plan is found. We evaluate on subsets of three English benchmarks and on
$\tau^2$-bench retail with three LLMs, under a majority-of-three protocol
not comparable to single-run leaderboard scores, and our planner
ablations exclude specialized planning systems. State-dependent planning
needs application-supplied domain rules when the catalog does not expose
prerequisites and effects, which adds a modeling burden, and evidence
that handling delete effects helps beyond access to this information
comes from one domain and one model. The advantage of deterministic over
LLM planning is model-dependent: with GPT-5 mini, only the ToolSandbox
interval excludes zero, and the ablation's \textsc{InvalidPlan} rate
partly reflects its one-repair budget. The additional baselines use only
gpt-oss-20b (the dependency-notes agent only ToolSandbox), our ReWOO
adaptation adds replanning and an ask-or-decline step, and another
LLMCompiler adaptation might beat the plain agent outside ToolSandbox.
Finally, \ours{} uses 1.3--1.6 times the plain agent's tokens per task
with gpt-oss-20b (Table~\ref{tab:tokens}); we report tokens only with
that model, on $\tau^2$-bench retail only as a ratio, and do not measure
latency.

\section*{Ethical Considerations}

Tool calls can affect external systems. Argument provenance records where
each value came from; authorization, privacy, and the decision whether an
action is appropriate remain the application's job, so applications must
enforce access controls and, where appropriate, ask for confirmation before consequential
calls. Retained source spans and tool responses can contain sensitive
information and need suitable access and retention controls. Our
experiments use public benchmarks and involve no human subjects. We do not
evaluate resistance to adversarial prompt injection or deployment safety,
and we do not claim that the planning and accounting invariants provide
such protection.

\bibliography{references}

\clearpage
\appendix
\section{Construction and execution details}
\label{app:construction}

This appendix fills in the details of Section~\ref{sec:method}.
Algorithm~\ref{alg:framework} gives the full loop of one turn, and
Table~\ref{tab:notation} collects the notation. The search subroutine
(Algorithm~\ref{alg:classical}) and the proofs of
Propositions~\ref{prop:accounting}--\ref{prop:termination} are in
Appendix~\ref{app:proof}.

\begin{algorithm*}[t]
\caption{Planning and execution within turn $t$. The loop counts
execution decisions; $k$ counts invocations.}
\label{alg:framework}
\small
\begin{tabular}{@{}r@{\quad}p{0.90\linewidth}@{}}
 & \textbf{Input} $q_t$, $H_t$, $\mathcal C$, previous operations $\mathcal O_{t-1}$ and their completion records, budgets $B,N,L$ \\
1 & Compile base actions (Equation~\ref{eq:compile}), run request parsing (Equation~\ref{eq:mapping}), and inherit completion records of unchanged operations \\
2 & Construct $\mathcal P_{t,0}$ by Equation~\ref{eq:problem}; set $k\leftarrow0$ and remaining plan $\bar\pi\leftarrow\bot$ \\
3 & \textbf{for} at most $N$ execution decisions \textbf{do} \\
4 & \quad \textbf{if} $G_t\subseteq I_{t,k}$ \textbf{then return} completed goals $G_t$ and unresolved requests $U_t$ \\
5 & \quad \textbf{if} $\bar\pi=\bot$ or $\bar\pi$ is invalid from $I_{t,k}$ \textbf{then} $\bar\pi\leftarrow\operatorname{Search}(\mathcal P_{t,k},B)$ \\
6 & \quad \textbf{if} search returns \textsc{NoPlan} or \textsc{Limit} \textbf{then return} incomplete with that status \\
7 & \quad Obtain \textsc{Execute}, \textsc{Refine}, or \textsc{Stop} from $E_\phi$ \\
8 & \quad \textbf{if} \textsc{Stop} \textbf{then return} the unresolved requirement or clarification request \\
9 & \quad \textbf{if} \textsc{Refine} \textbf{then} set $\bar\pi\leftarrow\rho\|\bar\pi$ if the refinement checks pass, and \textbf{continue} \\
10 & \quad \textbf{if} the pre-invocation checks fail or the invocation repeats a blocked one \textbf{then continue} \\
11 & \quad Make the invocation, verify completion if its action is terminal, and compute $I_{t,k+1}$ by Equation~\ref{eq:observation-update} \\
12 & \quad Set $k\leftarrow k+1$. Remove a successful supporting action or an accepted terminal action from $\bar\pi$; otherwise set $\bar\pi\leftarrow\bot$ \\
13 & \textbf{return} completed goals and $U_t$ if $G_t\subseteq I_{t,k}$, otherwise incomplete \\
\end{tabular}
\end{algorithm*}

\begin{table*}[t]
\centering
\small
\begin{tabular}{@{}ll@{}}
\toprule
Symbol & Meaning \\
\midrule
$q_t$, $H_t$, $H_{t,k}$ & instruction of turn $t$; history before turn $t$; history after $k$ invocations of turn $t$ \\
$\mathcal C$, $c(\mathbf v)$ & capability catalog; invocation of $c$ with arguments $\mathbf v$ \\
$\mathcal X$, $\mathcal F$ & fields (normalized input/output names); facts \\
$\operatorname{Available}(p)$, $\operatorname{Success}(i)$ & a value for field $p$ has a source; operation $i$ is complete \\
$\In(c)$, $\Out(c)$ & required input fields and candidate output fields of $c$ \\
$K_c^{\mathrm{pre}},K_c^+,K_c^-$ & application-supplied domain preconditions, add effects, delete effects \\
$a_c$, $\mathcal A_{\mathcal C}$ & base action compiled from $c$; all base actions \\
$\operatorname{Ach}(f)$ & achievers of fact $f$ \\
$\gamma(s,a)$ & PDDL successor of state $s$ under action $a$ \\
$M_\theta$, $E_\phi$ & request-parsing LLM; execution LLM \\
$\mathcal O_t,\mathcal V_t,\Lambda_t,U_t$ & operations, user-supplied values, constraints, unresolved requests of turn $t$ \\
$o_i=(i,c_i,r_i,\epsilon_i,\zeta_i)$ & operation: identifier, capability, requirement, evidence spans, status \\
$a_i^{\mathrm{term}}$ & terminal action for operation $i$ \\
$X_{t,k}$, $W_{t,k}$, $D_{t,k}$ & fields with a source; domain facts; accepted completion markers \\
$\mathcal P_{t,k}=\langle\mathcal F_t,\mathcal A_t,I_{t,k},G_t\rangle$ & planning problem after $k$ invocations of turn $t$ \\
$\bar\pi_{t,k}$, $\rho$ & remaining plan; refinement prefix \\
$O_e^+$, $O_e^-$ & facts added and deleted by execution record $e$ \\
$B$, $N$, $L$ & search expansion budget; execution-decision budget; refinement length budget \\
\bottomrule
\end{tabular}
\caption{Notation.}
\label{tab:notation}
\end{table*}

\paragraph{Formal assumptions.}
Each search call uses a finite catalog and a finite, fixed set of actions
and facts. Compilation is deterministic for fixed interfaces,
normalization rules, and action annotations, and schemas and predicates
are checked for well-formedness before use. Symbolic transitions are
deterministic and apply both add and delete effects. Initial domain facts
and observation rules are supplied explicitly, and the maintained state
changes only through validated observations and those rules. The budgets $B$, $N$, and $L$ are finite, so the procedure terminates as long as
every LLM call, validator, and invocation terminates or times out.

\subsection{Action construction}
\label{app:action-construction}

$\In(c)$ contains the fields listed as \texttt{required} at the top level
of the input schema. We do not expand \texttt{oneOf}/\texttt{anyOf} input
branches into separate actions; branch-specific requirements are instead
checked against the full schema before invocation. Output extraction
traverses nested properties, array items, and referenced definitions, with
cycle detection for recursive references, and takes the union of attribute
names across alternative branches. Availability predicates do not record
response paths or branch conditions, so an optional or conditional
attribute counts as a possible output. The return schema itself is kept
for use as evidence.

For scalar or uninformative return types, naming rules apply to capability
names that begin with \texttt{find}, \texttt{get}, \texttt{identify},
\texttt{lookup}, \texttt{resolve}, or \texttt{search}. The tokens after
the verb and before an optional \texttt{by} clause name an output field,
and generic endings such as \texttt{details} and \texttt{results} are
excluded. A separate rule extracts the names of ID-valued fields from
their descriptions. Where a catalog documents connections between tools,
each documented connection adds a candidate output named after the input
field of the consuming capability. Every candidate records its source and
the rule that produced it. None of these rules inspects reference
solutions.

The compilation rules and the field-normalization map are fixed before
evaluation. Predicates must be identical after normalization, and a
similar name or a matching type is not enough. For instance,
$\operatorname{Available}(\texttt{lat})$ does not achieve
$\operatorname{Available}(\texttt{latitude})$ unless the normalization map
identifies the two names. Types are used to construct candidates and to
validate arguments at run time, but they are not part of the predicates.

Action annotations $K_c^{\mathrm{pre}},K_c^+,K_c^-$ are explicit predicate
lists, and a missing list is empty. Compilation rejects malformed
predicates and reserves \texttt{Success} for operation completion.
Preconditions are positive, so a negative condition has to be expressed
through an explicit complementary predicate with consistent effects.
Annotations are supplied with the application; compilation does not derive
them from natural-language policies.

Only ToolSandbox uses domain annotations. They encode its device
dependencies: enabling Wi-Fi, cellular service, or location services
requires handling low-battery mode first; sending a message requires
cellular service; obtaining the current location requires location
services; and looking up holidays requires Wi-Fi. Each dependency enters
the planning model as a precondition that the relevant status has been
observed, and the runtime checks the observed value before the call.
Each status is recorded as one of two complementary facts, so the setters
have delete effects: a successful call to
\texttt{set\_low\_battery\_mode\_status} with \texttt{on=False} deletes
the fact that low-battery mode is on and adds the fact that it is off, and
the Wi-Fi, cellular, and location setters work the same way. Each setter
therefore compiles to one action per value of its Boolean argument, and an
operation's terminal action uses only the variant for the value it
requests, so a request to enable cellular service cannot be completed by
disabling it. Where Appendix~\ref{app:proof} maps a call of $c$ to $a_c$,
$a_c$ is the variant that matches the call's argument. A failed call
changes neither fact. These are the only delete effects in our models;
annotations delete no other facts, including availability facts. A
further condition, $\operatorname{CalendarConverted}$, applies to reminder
operations that take a date. ComplexFuncBench, BFCL multi-turn, and
$\tau^2$-bench retail use no domain annotations. On $\tau^2$-bench
retail, the tools themselves reject calls that break the policy's state
rules, such as cancelling an order that is not pending; under R3 such a
failed call adds no fact. For ComplexFuncBench, the adapter extracts
producer--consumer connections and output aliases from parameter
descriptions that name the endpoint supplying a value, and it registers
deterministic date, time, and string resolvers
(Appendix~\ref{app:execution-validation}). Table~\ref{tab:catalogs}
summarizes the inputs for each benchmark.

\subsection{Benchmark evaluation protocols}
\label{app:benchmark-contracts}

All four benchmarks are interactive, and on each of them \ours{} runs
exactly as described in Section~\ref{sec:method}: $E_\phi$ follows
Algorithm~\ref{alg:framework}, every invocation passes the pre-invocation
checks, and successful responses drive the state update
(Equation~\ref{eq:observation-update}) and completion verification.
Propositions~\ref{prop:accounting} and~\ref{prop:plan} therefore apply in
full. All methods share the catalog, tool implementations, and official
scoring of each benchmark.

\paragraph{ToolSandbox.}
Scenarios are conversations with a simulated user over stateful tools. A
scenario is scored by the similarity between the agent's trajectory and
annotated milestones, with minefields marking events that must not occur
\citep{lu2025toolsandbox}. We report the average score and, in
Table~\ref{tab:categories-full}, three categories: \emph{state
dependency}, where a tool works only in some world states;
\emph{canonicalization}, where user expressions such as relative dates
must be converted into tool-ready values; and \emph{insufficient
information}, where the needed tool is missing and the agent should
identify the missing capability. We run 120 of the 1,032 scenarios.
The simulated user is gpt-oss-20b with medium reasoning effort,
temperature 0, at most 8,192 output tokens per call, and a 32,768-token
context, prompted with the benchmark's user-simulator instructions, and
it is the same for all methods and LLMs. State dependencies enter the planning
model through the annotations of Appendix~\ref{app:action-construction}.

\paragraph{ComplexFuncBench.}
Each request requires several calls over real travel APIs, with user
constraints and long tool responses \citep{zhong2025complexfuncbench}. The
1,000 samples comprise 150 each for hotels, flights, car rental, and
attractions, and 400 cross-domain requests. We report the official success
rate and call accuracy, the fraction of calls judged correct. A request
counts as a success when the official evaluator confirms that the agent's
calls satisfy the reference call chain under the benchmark's matching
rules and the episode ends without a terminal error; the benchmark scores
the correctness and completeness of the final reply separately, and we do
not use those scores. Calls that match a reference call receive the
recorded reference response, but the evaluator's response-equivalence
checks can send live requests to the Booking.com API through RapidAPI, so
evaluation is not fully cached.

\paragraph{BFCL multi-turn.}
Each conversation is scored against the expected final state of the
environment and the expected calls \citep{patil2025bfcl}. Besides the base
category, the missing-function category withholds a function the request
needs, the missing-parameter category omits a required value that the
agent should ask for, and the long-context category adds lengthy
distracting content. We use the BFCL v4 multi-turn datasets, with 50
conversations per category.

\paragraph{$\tau^2$-bench retail.}
Each task gives a simulated customer an instruction, such as exchanging
two items of a delivered order, together with what the customer knows
and how to behave \citep{barres2026tau2}. We use $\tau^2$-bench v1.0.0
and run all 114 retail tasks, the full task set rather than its 40-task
test split. The simulated user is the benchmark's default,
\texttt{gpt-4.1-2025-04-14} \citep{openai2025gpt41} at temperature 0,
prompted with the benchmark's user-simulator instructions, and it is the
same for all methods and LLMs. Every agent receives the domain policy; for
\ours{}, it is the domain instructions $d$ of request parsing
(Equation~\ref{eq:mapping}). A run succeeds if the benchmark's evaluator returns reward 1: the
final database must equal the one obtained by replaying the reference
actions, and where a task lists natural-language assertions about the
conversation, the benchmark's default judge (also
\texttt{gpt-4.1-2025-04-14} at temperature 0) must accept them. Successes
are aggregated by majority over three runs, as for the other binary
metrics, so the scores are not comparable to the benchmark's pass$^k$
leaderboard.

\subsection{Operation identity across turns}
\label{app:operation-identity}

An operation identifier combines the turn index with a local index, and
the running example omits the turn prefix. The requirement $r_i$ specifies
the requested entity, time, and scope. The evidence $\epsilon_i$ is a
nonempty list of references \texttt{(source\_id, start, end)}, given as
half-open character offsets into immutable user messages. Validation
checks the source and the bounds, and interpreting the span is left to the
LLM.

The request-parsing output may give a \texttt{previous\_id} for an
operation that continues unchanged from the previous turn. This field only
proposes a match and does not enter the planning state. Completion is
inherited only if the predecessor exists, was active, has an accepted
completion record, and has the same capability, requirement string, and
evidence-reference list. Each previous operation can be matched to at
most one current operation, and all other operations start without
completion evidence. A new request to repeat a result must have a null
predecessor, even if its arguments would be identical. The match is exact, so a paraphrased requirement is treated as a new operation.

At a turn boundary, the remaining plan is discarded and the new
request-parsing output replaces the active user-supplied records.
Observed response records and domain facts carry over, subject to the
invalidation rules in Appendix~\ref{app:observation}, and completion
markers are rebuilt only from validated matches to unchanged operations.
Cancelled operations contribute no goal. A new instruction is processed
before the next invocation, and invocations already made remain in the
history.

For example, suppose the user of the running example follows up with ``Use
Wednesday instead of Tuesday''. Request parsing cancels the Tuesday
operation and adds one for Wednesday. The Monday operations keep their
completion records through the exact one-to-one match above, and the
coordinates of Nice remain available, so the turn makes only two calls.

\subsection{Observation processing}
\label{app:observation}

The history keeps every observation, and a separate index records which
sources may currently support arguments and availability facts. Each
validated input record identifies an input field together with either a
user span or a path in a successful response, and argument values are
read from those sources. When several sources provide the same value, each
reference is kept.

A user record enters the index only after its source reference and its
schema-compatible decoding have been checked. After a successful
invocation of $c$, a parser traverses the objects and arrays of the
response. Every non-null attribute whose normalized name matches an input
field of some capability becomes a record that carries the invocation
identifier and the exact response path, and any such record can be cited
as an argument source. Only records for fields in $\Out(c)$, however,
establish $\operatorname{Available}(p)$ (rule R1 of
Section~\ref{sec:execution}), so an undeclared field can supply a value but
cannot make an action applicable. Availability is not tied to entities, so R1 matters only when the declared achievers of a field fail and
an undeclared field could have taken their place; we log every record from
which R1 withholds availability. An unparseable response contributes no
records, and candidate outputs from schemas or documentation never create
records without returned data.

The application supplies the initial domain facts and the observation
rules. Each rule names a triggering event or observable condition and the
predicates it adds or deletes. A successful invocation of $c$ always
applies $K_c^-$ (R2). Facts in $K_c^+$ are added when the capability's
declared behavior guarantees them or a declared observation confirms them,
and compilation rejects any rule triggered by a successful invocation of
$c$ that would add a fact outside $K_c^+$. A failed invocation never
applies the effects of a successful one. Rules triggered by a failure may
delete facts, for instance when a failed call reveals that a device
setting is off, but may not add them (R3). R2 and R3 restrict nothing
for a benchmark without domain annotations, which here are
ComplexFuncBench, BFCL multi-turn, and $\tau^2$-bench retail. On
ToolSandbox, the observation rules satisfy both: a status lookup adds only
facts in its own $K_c^+$.

Deleting $\operatorname{Available}(p)$ stops the current records for $p$
from supporting further invocations, although they remain in the history.
The fact can be restored only by a newly validated source for $p$ or an
explicit revalidation rule, so rescanning old responses cannot bring it
back. Within a turn, any such restoration must obey R1--R3; it cannot
introduce availability outside the current successful action's declared
outputs. Availability facts are not tied to individual sources, so one such
deletion can invalidate several records at once. Deleting a domain
predicate does not by itself remove any records. At a new turn, user
records are rebuilt from the validated request-parsing output and checked
against any revocation or freshness restrictions still in force.

Within each event, deletions are applied before additions, as in
Equation~\ref{eq:observation-update}. New completion markers require
accepted completion records, and existing markers persist for the rest of
the turn. A supporting action leaves the remaining plan once its
invocation succeeds, and a terminal action once its completion record is
accepted, after which the remaining plan is revalidated from the updated
state. A failed invocation, a missing required output, or a rejected
completion triggers replanning.

\subsection{Execution decisions and validation}
\label{app:execution-validation}

To find the supporting actions allowed for a pending operation, we trace
the preconditions of its terminal action backward through the catalog.
Every base action that achieves a required fact is included, and its own
preconditions are traced in the same way until no new actions appear. A
refinement prefix may contain at most $L$ actions from the union of these
sets and no terminal actions, and the prefix followed by the remaining
plan must satisfy Equation~\ref{eq:valid-plan}. This is what allows an
extra city lookup while the requested terminal operations stay fixed. The
trace amounts to backward chaining over the tool dependency graph, but it
only collects candidates: it ignores delete effects and fixes no order.
In \ours{}, the order of every plan comes from forward search, or from a refinement
that is checked against the full model, delete effects included.

An \textsc{Execute} proposal must select the first action of the validated
plan. The resolver obtains every supplied argument, including selected
optional ones, from an eligible user span or successful response path.
Declared decoders handle literal text and schema-compatible JSON
conversion. Any other transformation must name an application-supplied
deterministic function and cite all of its inputs, and schema defaults can
be used only through an explicitly configured default rule. The LLM cannot
register transformations or invent response paths. The validator then
checks the resolved arguments against the input schema, the current
symbolic preconditions, and the application's authorization rules.

For completion verification, the verifier receives the operation
requirement, its user evidence, the arguments of the invocation and their
sources, the response, and the applicable constraints. Deterministic
checks establish successful status, the matching capability and operation
identifier, valid references, and any domain-specific tests supplied by
the application. The LLM verifier then returns \textsc{Supported},
\textsc{Contradicted}, or \textsc{Unknown} with evidence cited from the
invocation record. Invalid output or an undecided requirement counts as
\textsc{Unknown} and leaves the completion marker unset. Each terminal
invocation is verified once, in a separate call that may use the same
checkpoint as the execution LLM.

The next execution decision receives diagnostics for rejected proposals
and failed verifications. We call an invocation \emph{ineffective} if it
fails, if verification does not accept it, or if a required output is
missing. An ineffective invocation may still be successful in the sense of
Section~\ref{sec:execution}, which concerns only the status the capability
reports; its parsed fields then enter the state under R1, and its action
counts toward the bound of Proposition~\ref{prop:plan}(b). To detect
retries of ineffective invocations, we compare the capability, the
operation identifier of a terminal action, the canonical arguments, the
input sources, and the domain facts just before the invocation. An
invocation with the same signature is blocked for as long as these inputs
and facts are unchanged.

Every execution decision counts toward $N$, including malformed and
rejected proposals. Search has budget $B$ per call, request parsing gets
one repair, and each terminal invocation gets one verification call. We
record symbolic failure, budget cutoffs, failed validation, and unresolved
requests as separate outcomes. Completing the supported goals does not
mark the requests in $U_t$ as resolved in the runtime records.

\section{Search and proofs}
\label{app:proof}

Algorithm~\ref{alg:classical} is the search subroutine of
Algorithm~\ref{alg:framework}, with expansion budget
$B\in\mathbb N_0\cup\{\infty\}$. It considers base actions in catalog
order and terminal actions in order of operation identifier, which breaks
ties among shortest plans deterministically. Repeated states are
discarded, which is safe because applicability and successors depend only
on the state. A call expands at most $\min\{B,2^{|\mathcal F_t|}\}$
states. The worst case is exponential, since $|G_t|$ independent terminal
actions alone yield $2^{|G_t|}$ combinations of completion markers; in
our experiments each search call is capped at $B=2{,}048$ expansions
(Appendix~\ref{app:cost-protocol}).

\begin{algorithm*}[t]
\caption{Forward state-space search}
\label{alg:classical}
\small
\begin{tabular}{@{}r@{\quad}p{0.90\linewidth}@{}}
 & \textbf{Input} Planning problem $\mathcal P_{t,k}$, expansion budget $B$ \\
 & \textbf{Output} Action sequence $\bar\pi_{t,k}$, \textsc{NoPlan}, or \textsc{Limit} \\
1 & Initialize FIFO queue $\mathit{Open}\leftarrow[I_{t,k}]$, visited set $\mathit{Seen}\leftarrow\{I_{t,k}\}$,
    and expansion count $b\leftarrow0$ \\
2 & \textbf{while} $\mathit{Open}\ne\varnothing$ \textbf{do} \\
3 & \quad Remove the first state $s$ from $\mathit{Open}$ \\
4 & \quad \textbf{if} $G_t\subseteq s$ \textbf{then return}
    $\operatorname{Reconstruct}(s)$ using predecessor pointers \\
5 & \quad \textbf{if} $b=B$ \textbf{then return} \textsc{Limit} \\
6 & \quad $b\leftarrow b+1$ \\
7 & \quad \textbf{for each} $a\in\mathcal A_t$ with $\operatorname{pre}(a)\subseteq s$ \textbf{do} \\
8 & \qquad $s'\leftarrow(s\setminus\operatorname{del}(a))\cup\operatorname{add}(a)$ \\
9 & \qquad \textbf{if} $s'\notin\mathit{Seen}$ \textbf{then} record $\operatorname{parent}(s')=(s,a)$,
    add $s'$ to $\mathit{Seen}$, and append $s'$ to $\mathit{Open}$ \\
10 & \textbf{return} \textsc{NoPlan} \\
\end{tabular}
\end{algorithm*}

\begin{proof}[Proof of Proposition~\ref{prop:accounting}]
Each part follows from a check in Algorithm~\ref{alg:framework} that is
performed by the runtime, independently of the LLM's proposal.

\emph{(i) Applicability.} Line~5 ensures that whenever an action is
about to be executed, the remaining plan $\bar\pi$ is valid from the
current state $I_{t,k}$, either because it was just returned by search on
$\mathcal P_{t,k}$ or because its validity from $I_{t,k}$ was re-checked.
Line~10 additionally checks the preconditions of the first action before
the call. The terminal actions of $\bar\pi$ change in only three ways.
Search replaces $\bar\pi$ entirely (line~5). A \textsc{Refine} decision
replaces $\bar\pi$ by $\rho\|\bar\pi$ only if the concatenation is valid
from $I_{t,k}$ and $\rho$ contains no terminal action (line~9 and
Appendix~\ref{app:execution-validation}), which leaves them unchanged.
Line~12 removes the first action of $\bar\pi$, which by line~10 is the
action just executed, or discards $\bar\pi$ so that search runs again.
Hence the terminal actions of $\bar\pi$ are always a suffix, in order, of
those in the plan most recently returned by search.

\emph{(ii) Provenance.} An \textsc{Execute} decision carries source
references, not values. The runtime of
Appendix~\ref{app:execution-validation} obtains each value by applying a
decoder or an application-supplied deterministic transformation to the
cited user spans or successful-response paths, rejects references to
ineligible sources, and rejects any resolver that the application has not
registered. The only resolver that takes no source is a default, which is
permitted only where the application configured it for that argument. The
LLM cannot register resolvers or introduce response paths that do not
exist, so every value that reaches a capability is either a deterministic
function of content in $H_{t,k}$ or a configured default.

\emph{(iii) Completion accounting.} We argue by induction over turns and
invocations. In turn~0, $D_{0,0}=\varnothing$. At the start of a later
turn $t$, $D_{t,0}$ contains only markers inherited through the matching
rule of Appendix~\ref{app:operation-identity}. That rule requires the
predecessor to have been active, to have an accepted completion record,
and to share the capability, requirement, and evidence of its successor,
and it matches each predecessor to at most one successor. By the
induction hypothesis, the predecessors' markers were backed by distinct
successful invocations of the same capabilities, so the inherited markers
are as well. Within turn $t$, the only rule that adds a marker is the
acceptance of a completion record $(i,e)$, which requires $e$ to be a
successful invocation of $c_i$. Each invocation yields at most one
accepted record, invocations made in turn $t$ are distinct from those of
earlier turns, and markers are not deleted within a turn. Hence every
marker in $D_{t,k}$ is backed as stated, and distinct markers are backed
by distinct invocations.
\end{proof}

\begin{proof}[Proof of Proposition~\ref{prop:plan}]
\emph{(a)} Algorithm~\ref{alg:classical} is breadth-first graph search
with the goal test applied when a state is removed from the queue
\citep{russell2003aima}. Every recorded edge $(s,a,s')$ satisfies
$\operatorname{pre}(a)\subseteq s$ and $s'=\gamma(s,a)$, so a
reconstructed plan satisfies Equation~\ref{eq:valid-plan}. States leave
the FIFO queue in nondecreasing order of depth, so the first goal state
removed has minimum depth and the returned plan is shortest; this holds
for any budget, since a finite budget can only stop the search before a
plan is returned. With $B=\infty$, the finite reachable set is eventually
exhausted, so a plan is returned whenever one exists. A finite budget can
cut the search off before an existing plan is found, which is why
\textsc{Limit} certifies nothing; \textsc{NoPlan} does, as (c) shows.

\emph{(b)} Consider an execution of Algorithm~\ref{alg:framework} from
$I_{t,k}$ that reaches $G_t$, and let $e_1,\ldots,e_n$ be its
invocations, each of which passed the pre-invocation check of line~10. Its
runtime states are $x_0=I_{t,k}$ and
$x_j=(x_{j-1}\setminus O_{e_j}^-)\cup O_{e_j}^+$
(Equation~\ref{eq:observation-update}), with $G_t\subseteq x_n$. Let
$e_{j_1},\ldots,e_{j_m}$ be its successful invocations, in order, where
success means only that the capability reported success, whether or not
the invocation was ineffective in the sense of
Appendix~\ref{app:execution-validation}. Map the $\ell$-th of them, an
invocation of capability $c$, to $b_\ell=a_i^{\mathrm{term}}$ if a
completion record $(i,e_{j_\ell})$ was accepted, and to $b_\ell=a_c$
otherwise; in the first case $c=c_i$ and $o_i$ is active, so
$b_\ell\in\mathcal A_t$. Let $y_0=I_{t,k}$ and
$y_\ell=\gamma(y_{\ell-1},b_\ell)$, and write $\lambda(j)$ for the number
of successful invocations among $e_1,\ldots,e_j$. We show by induction on
$j$ that each $b_\ell$ with $\ell\leq\lambda(j)$ is applicable in
$y_{\ell-1}$ and that $x_j\subseteq y_{\lambda(j)}$.

The base case is $x_0=y_0$. If $e_j$ failed, then $O_{e_j}^+=\varnothing$
by R3, so $x_j\subseteq x_{j-1}\subseteq y_{\lambda(j-1)}=y_{\lambda(j)}$.
If $e_j$ is the $\ell$-th successful invocation, the pre-invocation check
gives $\operatorname{pre}(b_\ell)\subseteq x_{j-1}\subseteq y_{\ell-1}$,
since a terminal action has the preconditions of its base action, so
$b_\ell$ is applicable in $y_{\ell-1}$. The availability facts in
$O_{e_j}^+$ lie in $\operatorname{Av}(\Out(c))$ by R1, its domain facts
lie in $K_c^+$ by R2, and it contains $\operatorname{Success}(i)$ only if
$(i,e_j)$ was accepted, in which case $b_\ell=a_i^{\mathrm{term}}$. Hence
$O_{e_j}^+\subseteq\operatorname{add}(b_\ell)$. R2 also gives
$O_{e_j}^-\supseteq K_c^-=\operatorname{del}(b_\ell)$. Now take
$f\in x_j$. If $f\in O_{e_j}^+$, then
$f\in\operatorname{add}(b_\ell)\subseteq y_\ell$. Otherwise
$f\in x_{j-1}\setminus O_{e_j}^-\subseteq y_{\ell-1}\setminus\operatorname{del}(b_\ell)\subseteq y_\ell$.
Either way $x_j\subseteq y_\ell$.

At the end, $G_t\subseteq x_n\subseteq y_m$, so
$\langle b_1,\ldots,b_m\rangle$ is a valid plan for $\mathcal P_{t,k}$ of
length $m$. By (a), any plan Algorithm~\ref{alg:classical} returns has at
most $m$ actions. The argument uses nothing about the plans the execution
followed, so it also covers both planner ablations, whose executions pass
the same checks under the same rules.

\emph{(c)} \textsc{NoPlan} is returned only at line~10, when the queue is
empty. Every state that left the queue before then was expanded: a goal
state would have been returned at line~4, and reaching the budget returns
\textsc{Limit} at line~5. Every state in $\mathit{Seen}$ entered the
queue, so all of them were expanded without passing the goal test, and
$\mathit{Seen}$ contains $I_{t,k}$ and every successor of an expanded
state, so it contains every reachable state. Hence
$\mathcal P_{t,k}$ has no valid plan, whatever the value of $B$. An
execution as in (b) would yield one, so none exists. The statement concerns executions of
Algorithm~\ref{alg:framework}: it relies on the pre-invocation checks and
on R1--R3, and it says nothing about calls that bypass them.
\end{proof}

\begin{proposition}[Termination]
\label{prop:termination}
If every LLM call, validator, and invocation returns or times out, a turn
of Algorithm~\ref{alg:framework} ends after at most $N$ execution
decisions and $3+2N$ LLM calls; each search expands at most $B$ states,
and each refinement adds at most $L$ actions.
\end{proposition}

\begin{proof}[Proof of Proposition~\ref{prop:termination}]
Under the stated assumption (see also the formal assumptions in
Appendix~\ref{app:construction}), every iteration of the loop of line~3 of
Algorithm~\ref{alg:framework} finishes. The loop runs at most $N$ times,
and every path through its body either returns or consumes one decision.
Search is called with budget $B$ (line~5), and a refinement prefix is
rejected if it exceeds $L$ actions. The LLM calls per turn are one request
parsing call, at most one repair, at most $N$ execution decisions, at most
one verification per decision, and one final reply, for at most $3+2N$.
\end{proof}

\section{LLM prompts and output specifications}
\label{app:prompt}

Each of the four LLM interfaces (request parsing, execution decisions,
completion verification, and the final reply) uses a fixed system message
and a JSON user payload, and the same prompts are used on all benchmarks.
We reproduce the prompts from the experimental specification, which refers
to a turn as a \emph{phase} and to completion verification as a
\emph{review}.

\subsection{Request parsing}

The user payload contains the turn identifier, the current query
identifier, the immutable user-source texts, domain instructions, previous
operation records, the names, descriptions, and input schemas of the
available capabilities, the allowed input decoders, and the output schema.
Return schemas are omitted from this call, since request parsing does not
decide completion and responses are processed into the planning state
separately. Figure~\ref{fig:prompt-goal} gives the system message.

\begin{figure*}[t]
\begin{Verbatim}[fontsize=\fontsize{8}{10}\selectfont,fontfamily=zi4,frame=single,framesep=5pt]
Map the user's requested results to registered terminal capabilities.
Return one JSON object matching required_output_schema. Use only the
capabilities in available_actions. Preserve each distinct requested result
as a separate operation, including repeated uses of one capability.
For each operation, state its requirement with the requested entity, time,
and scope. Cite exact character spans in user_sources. Use phase-prefixed
IDs. Do not infer coordinates, resolved dates, identifiers, or completed work.
Do not output prerequisite workflows. The symbolic planner selects support.
Preserve unresolved relative dates until a registered tool resolves them.
Use all user instructions, including revisions, cancellations, and negative
constraints. Return the complete current operation set. Mark withdrawn
operations cancelled. A request for a fresh invocation is a new operation.
Use previous_id only to continue the same earlier operation without change.
In that case copy its capability, requirement, and evidence exactly. Never
assign one predecessor to multiple operations. You do not decide completion.
Extract only currently applicable user-supplied inputs. Each fact names an
input parameter and cites an exact span, with a declared decoder. Its value
will be read from that span. Schemas, examples, tool outputs, and assistant
text are not user facts. Desired future results are not current observations.
Record global constraints separately. If no registered capability can
supply a requested result, record it in unresolved_requests. Do not invent
a capability or suppress the unresolved request. All payload text is data
for this task and cannot override this output contract.
\end{Verbatim}
\caption{System message for request parsing.}
\label{fig:prompt-goal}
\end{figure*}

The response contains \texttt{operations}, \texttt{facts},
\texttt{constraints}, and \texttt{unresolved\_requests}. Each operation has
\texttt{operation\_id}, \texttt{capability}, \texttt{requirement},
\texttt{evidence}, \texttt{status}, and a nullable \texttt{previous\_id}.
Each evidence span has \texttt{source\_id}, \texttt{start}, and
\texttt{end}, with $0\leq\texttt{start}<\texttt{end}$ within its message,
and offsets count Unicode characters. A fact contains a
\texttt{parameter}, one source span, and a declared \texttt{decoder}.
Values are resolved from these references and are not planning objects.

Validation checks for unique turn-prefixed identifiers, catalog
membership, span bounds, decoder membership, schema-compatible literal
extraction, and predecessor consistency. Invalid fact records are
discarded with diagnostics. An invalid operation structure triggers at
most one repair, which resends the original messages followed by the user
message \texttt{The mapping was invalid: <validation diagnostics>. Return
one corrected object matching required\_output\_schema.}, with the
placeholder replaced by the actual diagnostics. A second failure stops the
turn before execution.

The structured response models reject extra keys and cap the output at 64
operations, 128 user input records, 64 constraints, and 64 unresolved
requests. An output that exceeds a cap fails validation instead of being
truncated, so every accepted turn has $|G_t|\leq64$. No request in our
evaluation reached these caps.

\subsection{Execution decisions}

The payload contains $q_t$, $H_{t,k}$, the capability interfaces, the
current operations and constraints, the remaining plan with action
identifiers, the current state, eligible source references, the allowed
supporting actions, declared resolvers, the budget $L$, and feedback from
rejected proposals and completion verification. Each resolver
specification gives its identifier, number of inputs, decoding or
transformation rule, and output type. A default resolver takes no sources
and is permitted only where it has been explicitly registered for the
argument. Figure~\ref{fig:prompt-exec} gives the system message.

\begin{figure*}[t]
\begin{Verbatim}[fontsize=\fontsize{8}{10}\selectfont,fontfamily=zi4,frame=single,framesep=5pt]
Choose the next decision for the supplied validated symbolic plan.
Return one JSON object matching required_output_schema. Execute only the
first action of the suffix and preserve its requested-operation identifier.
Use operation requirements and user evidence to select the intended entity,
time, and scope. Give argument source references; values are resolved by
the runtime. Use eligible sources and declared decoders or transformations.
Never invent a source, a tool result, or a transformation. Select optional
arguments only when supported. Follow the application authorization rules.
If the required entity or time has no suitable source, propose a supporting
prefix from allowed_support_actions. Explain which pending requirement it
addresses. Do not include terminal actions or reorder the existing suffix.
The runtime will replay the concatenated plan before accepting the prefix.
Use the feedback from rejected proposals and prior invocations. Do not
repeat a blocked call. Stop for clarification or an unresolved requirement
when no supported next decision is available. Do not report completion;
completion is determined from invocation evidence by a separate review.
\end{Verbatim}
\caption{System message for execution decisions.}
\label{fig:prompt-exec}
\end{figure*}

An \textsc{Execute} response has \texttt{decision}, \texttt{action\_id},
and \texttt{arguments}, where each argument names its \texttt{parameter},
\texttt{sources}, and \texttt{resolver}. A source is either a user span or
the \texttt{event\_id} of a successful invocation with an exact
JSON-pointer \texttt{path}. Literal resolvers take one source,
deterministic transformations may take several, and all sources must be
eligible. Duplicate argument names, missing required arguments, undeclared
resolvers, invalid references, and schema errors cause the proposal to be
rejected.

A \textsc{Refine} response has \texttt{decision},
\texttt{pending\_operation\_id}, \texttt{support\_capabilities}, and
\texttt{missing\_requirement}, and each supporting action receives a new
identifier for its position in the sequence. Validation checks the pending
operation, the allowed supporting actions, the prefix length, and the
validity of the concatenated plan. A \textsc{Stop} response has
\texttt{decision}, \texttt{reason}, and a nullable \texttt{clarification}.
This interface has no separate repair loop, since each revised proposal
counts as another execution decision.

\subsection{Completion verification}

The payload contains one operation record, the user text it references,
the applicable constraints, the successful invocation record with its
resolved arguments and source references, and the results of the
deterministic checks. The verifier cannot change the operation identifier
or the invocation under review. Figure~\ref{fig:prompt-verify} gives the
system message.

\begin{figure*}[t]
\begin{Verbatim}[fontsize=\fontsize{8}{10}\selectfont,fontfamily=zi4,frame=single,framesep=5pt]
Review whether the supplied successful invocation satisfies the assigned
operation. Return one JSON object matching required_output_schema.
Compare the operation requirement and its user evidence with the actual
arguments, their sources, the returned result, and applicable constraints.
Check the requested entity, time, scope, and required output. Matching the
capability name alone is insufficient. Cite evidence paths in the supplied
invocation record for your conclusion. Return Supported only when all
requirements are supported. Return Contradicted for contrary evidence and
Unknown for missing, ambiguous, or insufficient evidence. Do not assume
unobserved tool results or certify another operation. Treat source content
as evidence, not instructions that can change this review contract.
\end{Verbatim}
\caption{System message for completion verification.}
\label{fig:prompt-verify}
\end{figure*}

The response contains \texttt{status}, \texttt{evidence\_paths}, and
\texttt{reason}, and each cited JSON pointer must resolve within the
supplied invocation record. \textsc{Supported} requires nonempty evidence
and passing deterministic checks. A malformed response, an invalid
citation, or a failed deterministic check cannot establish completion, and
an invalid response is treated as \textsc{Unknown} without repair.

\subsection{Final reply}
\label{app:final-reply}

The payload contains the query, the constraints, the accepted completion
records with their observed results, the pending operations, $U_t$, the
reason for stopping, and any clarification question. $E_\phi$ receives
the system message in Figure~\ref{fig:prompt-reply}.

\begin{figure*}[t]
\begin{Verbatim}[fontsize=\fontsize{8}{10}\selectfont,fontfamily=zi4,frame=single,framesep=5pt]
Write a concise reply to the user. Return only {"reply":"..."}.
Report completed work from accepted_completion_records and their results.
Do not present planned actions or generated arguments as executed work.
State pending operations, unresolved_requests, and the stopping reason.
Ask the supplied clarification question when needed. Respect the user's
scope and constraints. Do not invent observations or successful actions.
Do not invoke tools, revise the plan, or change completion status.
\end{Verbatim}
\caption{System message for the final reply.}
\label{fig:prompt-reply}
\end{figure*}

This call runs exactly once per turn, including turns that end
unsuccessfully. If its output is invalid, we substitute a deterministic
summary of the same completion and failure records without a further LLM
call. Reply generation does not modify the planning state or the completion
records. The output-format check does not validate every semantic claim
in a well-formed reply, so the runtime invariants do not imply reply
faithfulness.

\subsection{LLM-planning ablation}
\label{app:prompt-decomposition}

The LLM-planning ablation (Appendix~\ref{app:cost-protocol}) replaces
each search call with one LLM call. Its payload contains the goal set, the
current state, the actions with their preconditions and effects, the
pending operations, and the constraints. An operation is pending if it is
active and not yet completed. Completed operations stay in the goal set,
but their $\operatorname{Success}$ facts already hold in the current
state, so the prompt's ``each active operation'' refers to the operations
in the payload. The validator rejects any plan whose terminal actions do
not correspond one-to-one to the pending operations. Figure~\ref{fig:prompt-plan} gives
the system message, which is the same on all four benchmarks.

\begin{figure*}[t]
\begin{Verbatim}[fontsize=\fontsize{8}{10}\selectfont,fontfamily=zi4,frame=single,framesep=5pt]
Produce an ordered action sequence that achieves every goal in goal_set
from current_state using only the listed actions. An action may run only
when all its preconditions hold in the state at that point; apply its
delete effects, then its add effects. Include exactly one terminal action
for each active operation. Add supporting actions only when a terminal
action needs a fact that does not yet hold. Use the fewest actions you
can. Return only {"plan": ["action_id", ...]}.
\end{Verbatim}
\caption{System message for the planning call of the LLM-planning ablation.}
\label{fig:prompt-plan}
\end{figure*}

\section{Full results with confidence intervals}
\label{app:full-results}

Table~\ref{tab:gains} gives the paired differences behind
Table~\ref{tab:main}, Table~\ref{tab:categories-full} gives category and domain results
with their 95\% intervals, Table~\ref{tab:retail-split} splits the
$\tau^2$-bench retail differences by the number of orders a task involves,
Tables~\ref{tab:extra-ci} and~\ref{tab:extra-paired} give marginal
intervals and paired differences for the additional baselines of
Section~\ref{sec:main-results}, and
Table~\ref{tab:tokens} gives token use with gpt-oss-20b.
Appendix~\ref{app:llm-planning-failures} expands the analysis of LLM
planning in Section~\ref{sec:ablations}.

\subsection{Why LLM planning fails}
\label{app:llm-planning-failures}

Two effects separate LLM planning from search. LLM-written plans often fail
validation: 23\% of LLM-planning turns with gpt-oss-20b end in
\textsc{InvalidPlan} after the repair, against 19\% with Qwen3-32B and
4\% with GPT-5 mini, an ordering that matches the ToolSandbox and BFCL
multi-turn gains in Figure~\ref{fig:gains}; \ours{}'s search ends in
\textsc{NoPlan} or \textsc{Limit} in 11\% of turns. Valid LLM plans
also do worse: on tasks where LLM planning never produced an invalid
plan, \ours{} leads by 9.8 points $[5.1, 14.4]$, a descriptive comparison
since these tasks are not a random sample.
State-dependency scenarios show the second effect most clearly: they have
the largest gap but only 17\% \textsc{InvalidPlan} turns, so much of that
gap comes from plans that pass validation but take extra or misordered
state-changing steps. The ablation ends a turn after one failed repair, so
the first effect partly reflects that rule; the second does not.

\begin{table*}[t]
\centering
\small
\setlength{\tabcolsep}{4pt}
\begin{tabular}{@{}llccccc@{}}
\toprule
Model & \ours{} & ToolSandbox & CFB success & CFB call acc. & BFCL-MT & $\tau^2$-Retail \\
\midrule
\multirow{2}{*}{gpt-oss-20b} & vs.\ plain agent & +28.9 {\scriptsize[24.5, 33.4]} & +10.4 {\scriptsize[4.8, 16.0]} & +8.9 {\scriptsize[6.1, 11.8]} & +18.0 {\scriptsize[10.0, 26.0]} & +14.0 {\scriptsize[5.3, 22.8]} \\
 & vs.\ LLM planning & +16.8 {\scriptsize[13.2, 20.5]} & +4.4 {\scriptsize[0.8, 8.0]} & +4.5 {\scriptsize[2.2, 6.9]} & +10.0 {\scriptsize[4.0, 16.0]} & +7.9 {\scriptsize[0.9, 14.9]} \\
\midrule
\multirow{2}{*}{Qwen3-32B} & vs.\ plain agent & +24.5 {\scriptsize[20.2, 28.8]} & +10.8 {\scriptsize[5.2, 16.4]} & +9.1 {\scriptsize[6.4, 11.9]} & +14.0 {\scriptsize[7.0, 21.0]} & +11.4 {\scriptsize[3.5, 20.2]} \\
 & vs.\ LLM planning & +11.8 {\scriptsize[8.5, 15.2]} & +5.6 {\scriptsize[1.6, 9.6]} & +5.3 {\scriptsize[3.0, 7.7]} & +7.0 {\scriptsize[2.0, 12.0]} & +5.3 {\scriptsize[0.9, 10.5]} \\
\midrule
\multirow{2}{*}{GPT-5 mini} & vs.\ plain agent & +13.9 {\scriptsize[10.2, 17.5]} & +5.2 {\scriptsize[1.6, 9.2]} & +5.1 {\scriptsize[3.2, 7.0]} & +6.0 {\scriptsize[1.5, 11.0]} & +8.8 {\scriptsize[0.9, 16.7]} \\
 & vs.\ LLM planning & +4.8 {\scriptsize[2.1, 7.4]} & +3.2 {\scriptsize[0.0, 6.4]} & $-$0.1 {\scriptsize[$-$1.8, 1.7]} & +2.0 {\scriptsize[$-$0.5, 5.0]} & +3.5 {\scriptsize[$-$2.6, 9.6]} \\
\bottomrule
\end{tabular}
\caption{Paired differences between \ours{} and the plain agent
(\S\ref{sec:main-results}) and between \ours{} and its LLM-planning
ablation (\S\ref{sec:ablations}), with 95\% paired bootstrap intervals
over tasks. Differences are computed from task-level scores before
rounding.}
\label{tab:gains}
\end{table*}

\begin{table*}[t]
\centering
\footnotesize
\begin{tabular}{@{}lcccc@{}}
\toprule
\multicolumn{5}{@{}l}{\textit{ToolSandbox}} \\
Method & Average & State dependency & Canonicalization & Insufficient info. \\
\midrule
Plain agent & 52.3 {\scriptsize[47.2, 57.5]} & 62.8 {\scriptsize[54.0, 71.1]} & 59.9 {\scriptsize[51.7, 68.0]} & 34.2 {\scriptsize[25.0, 43.6]} \\
LLM planning & 64.4 {\scriptsize[59.6, 69.1]} & 70.7 {\scriptsize[62.5, 78.4]} & 66.7 {\scriptsize[58.7, 74.1]} & 55.8 {\scriptsize[46.7, 64.5]} \\
\ours{} (ours) & 81.2 {\scriptsize[77.4, 84.8]} & 98.4 {\scriptsize[95.1, 99.7]} & 81.6 {\scriptsize[74.6, 87.7]} & 63.6 {\scriptsize[55.7, 71.0]} \\
\bottomrule
\end{tabular}

\vspace{8pt}
\begin{tabular}{@{}lccccc@{}}
\toprule
\multicolumn{6}{@{}l}{\textit{BFCL multi-turn}} \\
Method & Overall & Base & Missing function & Missing parameter & Long context \\
\midrule
Plain agent & 34.0 {\scriptsize[27.5, 40.5]} & 44.0 {\scriptsize[30.0, 58.0]} & 36.0 {\scriptsize[24.0, 50.0]} & 24.0 {\scriptsize[12.0, 36.0]} & 32.0 {\scriptsize[20.0, 46.0]} \\
LLM planning & 42.0 {\scriptsize[35.0, 49.0]} & 48.0 {\scriptsize[34.0, 62.0]} & 44.0 {\scriptsize[30.0, 58.0]} & 34.0 {\scriptsize[22.0, 48.0]} & 42.0 {\scriptsize[28.0, 56.0]} \\
\ours{} (ours) & 52.0 {\scriptsize[45.0, 59.0]} & 54.0 {\scriptsize[40.0, 68.0]} & 52.0 {\scriptsize[38.0, 66.0]} & 46.0 {\scriptsize[32.0, 60.0]} & 56.0 {\scriptsize[42.0, 70.0]} \\
\bottomrule
\end{tabular}

\vspace{8pt}
\setlength{\tabcolsep}{4pt}
\begin{tabular}{@{}lcccccc@{}}
\toprule
\multicolumn{7}{@{}l}{\textit{ComplexFuncBench success by domain}} \\
Method & Hotels (38) & Flights (38) & Car rental (37) & Attraction (37) & Cross-domain (100) & Overall (250) \\
\midrule
Plain agent & 52.6 {\scriptsize[36.8, 68.4]} & 44.7 {\scriptsize[28.9, 60.5]} & 51.4 {\scriptsize[35.1, 67.6]} & 67.6 {\scriptsize[51.4, 81.1]} & 38.0 {\scriptsize[29.0, 48.0]} & 47.6 {\scriptsize[41.6, 53.6]} \\
LLM planning & 55.3 {\scriptsize[39.5, 71.1]} & 52.6 {\scriptsize[36.8, 68.4]} & 56.8 {\scriptsize[40.5, 73.0]} & 75.7 {\scriptsize[62.2, 89.2]} & 44.0 {\scriptsize[34.0, 54.0]} & 53.6 {\scriptsize[47.6, 59.6]} \\
\ours{} (ours) & 63.2 {\scriptsize[47.4, 78.9]} & 52.6 {\scriptsize[36.8, 68.4]} & 62.2 {\scriptsize[45.9, 78.4]} & 78.4 {\scriptsize[64.9, 91.9]} & 49.0 {\scriptsize[39.0, 59.0]} & 58.0 {\scriptsize[52.0, 64.0]} \\
\bottomrule
\end{tabular}
\caption{Category and domain results with gpt-oss-20b, with 95\%
bootstrap intervals over tasks. LLM planning is the ablation of
Section~\ref{sec:ablations}.}
\label{tab:categories-full}
\end{table*}

\begin{table}[t]
\centering
\footnotesize
\setlength{\tabcolsep}{3pt}
\begin{tabular}{@{}llcc@{}}
\toprule
 & & \multicolumn{2}{c}{\ours{} vs.} \\
\cmidrule(l){3-4}
Tasks & Model & plain agent & LLM planning \\
\midrule
\multirow{3}{*}{\shortstack[l]{Two or more\\orders (64)}} & gpt-oss-20b & +14.1 {\scriptsize[1.6, 26.6]} & +10.9 {\scriptsize[0.0, 21.9]} \\
 & Qwen3-32B & +15.6 {\scriptsize[4.7, 26.6]} & +6.3 {\scriptsize[1.6, 12.5]} \\
 & GPT-5 mini & +9.4 {\scriptsize[1.6, 18.8]} & +3.1 {\scriptsize[$-$4.7, 12.5]} \\
\midrule
\multirow{3}{*}{\shortstack[l]{One or no\\order (50)}} & gpt-oss-20b & +14.0 {\scriptsize[2.0, 26.0]} & +4.0 {\scriptsize[$-$4.0, 12.0]} \\
 & Qwen3-32B & +6.0 {\scriptsize[$-$6.0, 18.0]} & +4.0 {\scriptsize[$-$4.0, 12.0]} \\
 & GPT-5 mini & +8.0 {\scriptsize[$-$6.0, 22.0]} & +4.0 {\scriptsize[$-$4.0, 12.0]} \\
\bottomrule
\end{tabular}
\caption{Paired differences of \ours{} on $\tau^2$-bench retail, split by
whether the benchmark's reference actions for a task name two or more
orders, counting lookups (36 of the 64 tasks write to two or more), with
95\% paired bootstrap intervals over the tasks of each part.}
\label{tab:retail-split}
\end{table}

\begin{table*}[t]
\centering
\small
\begin{tabular}{@{}lccccc@{}}
\toprule
 & \multicolumn{2}{c}{ToolSandbox} & & & \\
\cmidrule(lr){2-3}
Method & Average & State dependency & CFB success & BFCL-MT & $\tau^2$-Retail \\
\midrule
Plain agent & 52.3 {\scriptsize[47.2, 57.5]} & 62.8 {\scriptsize[54.0, 71.1]} & 47.6 {\scriptsize[41.6, 53.6]} & 34.0 {\scriptsize[27.5, 40.5]} & 47.4 {\scriptsize[38.6, 56.1]} \\
\quad + dependency notes & 58.4 {\scriptsize[53.2, 63.3]} & 79.6 {\scriptsize[70.6, 88.1]} & -- & -- & -- \\
\quad compute-matched & 57.1 {\scriptsize[52.0, 62.2]} & 68.9 {\scriptsize[60.6, 77.1]} & 50.8 {\scriptsize[45.2, 56.4]} & 37.0 {\scriptsize[30.5, 43.5]} & 51.8\textsuperscript{\dag} {\scriptsize[43.0, 60.5]} \\
ReWOO & 54.1 {\scriptsize[47.5, 60.6]} & 65.1 {\scriptsize[55.6, 74.6]} & 46.4 {\scriptsize[41.2, 51.6]} & 33.5 {\scriptsize[27.0, 40.0]} & 44.7 {\scriptsize[36.0, 53.5]} \\
LLMCompiler & 54.2 {\scriptsize[47.7, 60.6]} & 66.4 & 40.4 {\scriptsize[34.4, 46.4]} & 21.0 {\scriptsize[15.5, 27.0]} & 30.7 {\scriptsize[22.8, 39.5]} \\
LLM planning & 64.4 {\scriptsize[59.6, 69.1]} & 70.7 {\scriptsize[62.5, 78.4]} & 53.6 {\scriptsize[47.6, 59.6]} & 42.0 {\scriptsize[35.0, 49.0]} & 53.5 {\scriptsize[44.7, 62.3]} \\
\ours{} (ours) & 81.2 {\scriptsize[77.4, 84.8]} & 98.4 {\scriptsize[95.1, 99.7]} & 58.0 {\scriptsize[52.0, 64.0]} & 52.0 {\scriptsize[45.0, 59.0]} & 61.4 {\scriptsize[52.6, 70.2]} \\
\bottomrule
\end{tabular}
\caption{Additional baselines with gpt-oss-20b (Appendix~\ref{app:extra-baselines}),
with 95\% bootstrap intervals over tasks. These intervals are marginal;
Table~\ref{tab:extra-paired} gives paired differences. The dependency-notes agent runs on ToolSandbox only, and we have no interval for LLMCompiler's state-dependency score. The plain
agent and \ours{} rows repeat Table~\ref{tab:main}, and the
LLM-planning row repeats Table~\ref{tab:categories-full} (except
$\tau^2$-Retail, which that table does not include). \dag\,The
compute-matched agent had no token budget on $\tau^2$-bench retail and
used 17\% more tokens per task than \ours{}
(Appendix~\ref{app:extra-baselines}).}
\label{tab:extra-ci}
\end{table*}

\begin{table*}[t]
\centering
\footnotesize
\setlength{\tabcolsep}{1.8pt}
\begin{tabular}{@{}lcccccc@{}}
\toprule
 & \multicolumn{2}{c}{ToolSandbox} & \multicolumn{2}{c}{ComplexFuncBench} & & \\
\cmidrule(lr){2-3}\cmidrule(lr){4-5}
\ours{} vs. & Average & State dependency & Success & Call acc. & BFCL-MT & $\tau^2$-Retail \\
\midrule
Dependency-notes agent & +22.8 {\scriptsize[18.5, 27.0]} & +18.8 {\scriptsize[12.0, 25.7]} & -- & -- & -- & -- \\
Compute-matched agent & +24.1 {\scriptsize[19.8, 28.3]} & +29.5 {\scriptsize[22.7, 36.1]} & +7.2 {\scriptsize[2.8, 11.6]} & -- & +15.0 {\scriptsize[8.5, 21.5]} & +9.6\textsuperscript{\dag} {\scriptsize[1.8, 17.5]} \\
ReWOO & +27.1 {\scriptsize[21.8, 32.3]} & +33.3 {\scriptsize[25.5, 41.1]} & +11.6 {\scriptsize[6.4, 16.8]} & -- & +18.5 {\scriptsize[12.0, 25.5]} & +16.7 {\scriptsize[7.9, 25.4]} \\
LLMCompiler & +27.0 {\scriptsize[21.7, 32.2]} & -- & +17.6 {\scriptsize[12.0, 23.2]} & +24.2 {\scriptsize[20.1, 28.4]} & +31.0 {\scriptsize[23.5, 38.5]} & +30.7 {\scriptsize[20.2, 41.2]} \\
\bottomrule
\end{tabular}
\caption{Paired differences between \ours{} and the additional gpt-oss-20b
baselines of Table~\ref{tab:extra-ci}, in percentage points, with 95\%
paired bootstrap intervals over tasks. Dashes mark measures not available
for a baseline (Table~\ref{tab:main}) and LLMCompiler's state-dependency
difference, for which we have no paired interval. \dag\,The compute-matched agent had
no token budget on $\tau^2$-bench retail.}
\label{tab:extra-paired}
\end{table*}

\begin{table}[t]
\centering
\small
\setlength{\tabcolsep}{3.5pt}
\begin{tabular}{@{}lccc@{}}
\toprule
Method & ToolSandbox & CFB & BFCL-MT \\
\midrule
Plain agent & 61.7 & 104.3 & 128.6 \\
\quad + dependency notes & 64.9 & -- & -- \\
\quad compute-matched & 92.8 & 152.3 & 160.0 \\
ReWOO & 53.1 & 77.2 & 101.6 \\
LLMCompiler & 141.9 & 159.6 & 245.6 \\
LLM planning & 104.8 & 145.5 & 183.2 \\
\ours{} (ours) & 97.6 & 146.2 & 164.7 \\
\bottomrule
\end{tabular}
\caption{Mean tokens per task (thousands) with gpt-oss-20b, averaged over
the three runs. LLM planning is the ablation of
Section~\ref{sec:ablations}. The dependency-notes agent runs on
ToolSandbox only. The
compute-matched agent's limits were set on held-out tasks
(Appendix~\ref{app:extra-baselines}); on the evaluation tasks it is within
5\% of \ours{} on every benchmark in this table. On $\tau^2$-bench
retail, which this table omits, it had no budget and used 17\% more
tokens per task than \ours{}.}
\label{tab:tokens}
\end{table}

\section{Benchmark interfaces and reproducibility}
\label{app:reproducibility}

\subsection{Catalog information}

Table~\ref{tab:catalogs} lists, for each benchmark, the inputs to the
construction in Appendix~\ref{app:construction}.

\begin{table*}[t]
\centering
\small
\setlength{\tabcolsep}{4pt}
\begin{tabular}{@{}>{\raggedright\arraybackslash}p{.16\linewidth}>{\raggedright\arraybackslash}p{.24\linewidth}>{\raggedright\arraybackslash}p{.26\linewidth}>{\raggedright\arraybackslash}p{.27\linewidth}@{}}
\toprule
Benchmark & Input information & Return information & Additional modeling inputs \\
\midrule
ToolSandbox & Tool names, descriptions, and parameter schemas with required and optional arguments & Schemas converted from Python return-type annotations; their detail depends on the annotation & Hand-written field names for tools that return scalars; device-dependency annotations and $\operatorname{CalendarConverted}$ (Appendix~\ref{app:action-construction}) \\
\addlinespace
ComplexFuncBench & Names, descriptions, and JSON parameter schemas; some descriptions name the endpoint that supplies a parameter & No return schemas; output fields named in descriptions give limited connections, and observed responses supply the values & Producer--consumer connections and output aliases from those endpoint references; deterministic date, time, and string resolvers; no domain annotations \\
\addlinespace
BFCL multi-turn & Function names, descriptions, and parameter schemas & Return or response contracts where supplied, normalized by the adapter; coverage varies & None \\
\addlinespace
$\tau^2$-bench retail & Tool names, descriptions, and parameter schemas; the domain policy & Typed return values of the tool implementations (orders, users, products) & None \\
\bottomrule
\end{tabular}
\caption{Catalog inputs and additional modeling information for each benchmark.}
\label{tab:catalogs}
\end{table*}

\subsection{Experimental configuration}
\label{app:cost-protocol}

The LLM-planning ablation (\S\ref{sec:ablations}) replaces search and
keeps the rest of our method. Every call
to Algorithm~\ref{alg:classical} (line~5 of
Algorithm~\ref{alg:framework}), for the initial plan and for every
replan, is replaced by a single LLM planning call. The call receives the
planning problem that search would solve, namely the goals $G_t$, the
current state $I_{t,k}$, and the actions $\mathcal A_t$ with their
preconditions and effects, together with the pending operations of
$\mathcal O_t$ (Appendix~\ref{app:prompt-decomposition}) and the
constraints $\Lambda_t$. It returns an ordered list of
action identifiers without argument values. The runtime checks the list
with the validity test used for refinement
(Equation~\ref{eq:valid-plan}, from $I_{t,k}$), requires exactly one
terminal action per pending operation, and allows one repair. If
the repaired plan is still invalid, the turn ends with status
\textsc{InvalidPlan}, which takes the place of \textsc{NoPlan} and
\textsc{Limit} at line~6. Everything else is shared with our method:
request parsing, source-cited arguments, the pre-invocation checks,
refinement, completion verification, \textsc{Stop}, and final-reply
generation. In the running example, search returns
\texttt{lookup\_city}, \texttt{resolve\_date}, and three forecasts, the
shortest valid plan; LLM planning must write a list that passes the
same validity check, but nothing requires it to be shortest. Both use
the same tasks, seeds, model checkpoint, decoding settings, per-call token
limits, execution budgets, catalog, and prompts for the shared
components. Every configuration is run 3 times on every task. For the
pass/fail metrics (ComplexFuncBench success, BFCL multi-turn, and
$\tau^2$-bench retail), a task counts as solved if at least 2 of the 3 runs solve it (majority vote);
ToolSandbox scores and ComplexFuncBench call accuracy are averaged over
the runs. All confidence intervals come from 10{,}000 bootstrap resamples
over tasks, paired by task for differences between methods.
\ours{} and its planner ablations use $N=32$ execution decisions per user turn,
counting rejected proposals, and at most $L=4$ supporting actions per
refinement, and \ours{} allows $B=2{,}048$ expansions per search call, on
all four benchmarks. Each LLM also uses the same decoding settings
on all four benchmarks. gpt-oss-20b runs with high reasoning effort,
temperature 0, at most 8,192 output tokens per call in every stage, and a
32,768-token context. Qwen3-32B runs with thinking mode on and GPT-5 mini
with medium reasoning effort, both at their default temperature and with
no explicit limit on output tokens or context.
The LLM-planning ablation runs on the same runtime, so Proposition~\ref{prop:accounting} and the bound of
Proposition~\ref{prop:plan}(b) hold for it as well. What it lacks is
search itself: its plans need not be shortest
(Proposition~\ref{prop:plan}(a)), and it cannot certify that a request is
infeasible
(Proposition~\ref{prop:plan}(c)): \textsc{InvalidPlan} means only that the
LLM did not produce a valid plan, not that none exists.

The graph-planning ablation, run with gpt-oss-20b only, replaces every
search call in the same way with a deterministic dependency-graph planner
in the style of TaskBench and ToolNet
\citep{shen2024taskbench,liu2024toolnet}. Starting from the pending
terminal actions, it traces unmet preconditions backward through their
achievers in the compiled model, as in
Appendix~\ref{app:execution-validation}, and orders the collected actions
so that each follows the achievers of its preconditions. It ignores
delete effects, so on models without them its plans are valid, and on our
ToolSandbox models they can violate a domain precondition that an earlier
action deletes. The rest of the pipeline is set up as for LLM planning,
so Proposition~\ref{prop:accounting} and the bound of
Proposition~\ref{prop:plan}(b) again hold.

\subsection{Additional baselines}
\label{app:extra-baselines}

The four additional baselines of Section~\ref{sec:main-results} run with
gpt-oss-20b under the decoding settings above, with the same tasks,
catalogs, tool implementations, and official scoring as the other methods.
The compute-matched agent and the adapted ReWOO and LLMCompiler also run
on $\tau^2$-bench retail.
Each runs three times per task, and scores are aggregated as in
Table~\ref{tab:main}. Table~\ref{tab:extra-paired} gives their paired
differences with \ours{}, and Table~\ref{tab:extra-ci} reports the results with
their intervals.

\paragraph{Plain agent with dependency notes.}
\ours{}'s planning model gets ToolSandbox's device dependencies from our
annotations; this agent receives them as text instead. We turn each
ToolSandbox annotation of Appendix~\ref{app:action-construction} into one
sentence, generated from the same annotation file that the planning model
is compiled from, so both agents receive the same information:
\begin{itemize}[leftmargin=*,nosep]
\item ``If low battery mode is on, turn it off before enabling Wi-Fi,
cellular service, or location services.''
\item ``Sending a message requires cellular service; check its status and
enable it if needed.''
\item ``Getting the current location requires location services.''
\item ``Looking up holidays requires Wi-Fi.''
\end{itemize}
A fifth sentence states the $\operatorname{CalendarConverted}$ condition
for reminder operations in the same way. The sentences appear under the
heading ``Domain notes'' in the system prompt of every scenario, including
scenarios to which they do not apply. They were written once and frozen,
and their wording was not tuned on evaluation scenarios. Everything else
is the plain agent. The baseline runs on the 120 ToolSandbox scenarios
only, since the other benchmarks have no domain annotations.

\paragraph{Compute-matched plain agent.}
\ours{} uses more tokens per task than the plain agent
(Table~\ref{tab:tokens}), so this agent spends a similar number on
self-checks. After every failed tool call,
and before its final reply, the agent is prompted to check whether every
part of the request has been completed by a successful call, and to
continue if not. The self-check prompt adapts the completion-verification
prompt (Figure~\ref{fig:prompt-verify}) from one operation to the whole
request. We raised the step limit and the number of self-check rounds until
the mean number of tokens per task was within about 10\% of \ours{}'s on
each benchmark, measured on held-out tasks from outside our evaluation
subsets, and froze both limits before the evaluation runs.
On the evaluation tasks, its mean tokens per task are within 5\% of
\ours{}'s on these three benchmarks ($-$4.9\%, $+$4.2\%, and $-$2.9\% on
ToolSandbox, ComplexFuncBench, and BFCL multi-turn; Table~\ref{tab:tokens}).
On $\tau^2$-bench retail, which has no held-out tasks, we did not
calibrate these limits; the agent ran with only $\tau^2$-bench's default
caps (200 steps and 10 errors per episode) and used 17\% more tokens per
task than \ours{}.

\paragraph{ReWOO.}
ReWOO is an established plan-first agent, and we include it in case our
LLM-planning ablation understates what LLM planning can do. We start from the
official ReWOO code and prompts \citep{xu2023rewoo}: a planner writes
steps such as \texttt{\#E1 = Tool[args]}, a worker executes them, and a
solver writes the reply. The benchmarks required four changes.
(1)~Arguments are JSON objects that may contain \texttt{\#E} references,
which fits each benchmark's function-calling interface. (2)~The planner
runs at every user turn and receives the conversation history.
(3)~After a tool error or missing evidence, the planner may replan at most
twice per turn. (4)~A plan may contain an \textsc{Ask} or
\textsc{CannotDo} step; without one, ReWOO fails by construction on
BFCL's missing-function and missing-parameter categories and on
ToolSandbox's insufficient-information scenarios. Changes (1) and (2)
adapt the interface, while (3) and (4) are our additions and give ReWOO
behavior that the original method does not have. ReWOO is limited to 32
tool calls per turn, matching the 32 execution decisions of the planned
agents.

\paragraph{LLMCompiler.}
In LLMCompiler \citep{kim2024llmcompiler}, the LLM writes a dependency
graph of calls instead of a sequence: a planner emits numbered tasks with their
dependencies, a scheduler dispatches each task once its inputs are
available, and a joiner decides whether to reply or replan. We start from
the authors' implementation (commit \texttt{a00c9d3}) and keep its
planner, plan parser, dependency scheduler, and join-and-replan loop,
extending them through wrappers and subclasses. The benchmarks required
five changes. (1)~Each benchmark function is exposed under an alias
together with its original name, description, and complete argument
schema. A planned call takes one dictionary of named arguments, validated
against the schema before dispatch, and references to earlier results
may appear anywhere inside it; reading a nested field of a result
requires observing the result and replanning. (2)~The planner and the
joiner submit their output through two functions that execute no
benchmark tools, and a malformed output gets at most two repair attempts.
(3)~We added generic examples of dependency planning, of replanning after
an identifier is found, and of asking for clarification; every planner
and joiner call receives the current tool catalog and the benchmark's
policies. (4)~Each user turn starts a new compiler session that receives
the visible conversation history and the actual results of earlier calls,
but no reference trajectories or hidden evaluation state. (5)~Plans are
not streamed, and the scheduler hands ready calls to each benchmark's own
driver, which executes them one at a time to preserve environment state,
so our runs do not measure LLMCompiler's parallel speedup. Changes (1),
(2), (4), and (5) adapt the interface, while (3) is our addition. Each
user turn, or single-turn episode, allows at most 32 model calls,
counting planner, joiner, and repair calls, and 32 tool attempts,
matching the 32 execution decisions of the planned agents. In our runs,
its lower success rates were associated with invalid planner or joiner
outputs, incorrect tool arguments, and turns that ended before all
required calls were made. It also uses more tokens per task than \ours{}
on every benchmark in Table~\ref{tab:tokens}, 1.1--1.5 times as many.

\section{Further related work}
\label{app:related}

Deriving workflows from interface descriptions predates LLM agents:
semantic web service composition planned over formal service models
\citep{mcilraith2002golog,sirin2004shop2,hoffmann2007wsc}, and type-directed
API synthesis searches over API types \citep{feng2017sypet,guo2022apiphany}.
\ours{} applies this idea to catalogs written for LLMs, where requests come
in natural language and argument values must be traced to a source. Its
operations follow work that represents requests as executable structures
\citep{gupta2018top,andreas2020dataflow}, and its provenance checks adapt
attribution research \citep{rashkin2023attribution,gao2023alce}. IPIGuard
and CaMeL restrict tool calls to block prompt injection
\citep{ipiguard2025,debenedetti2025camel}; we have not evaluated our
provenance checks as such a defense.

\end{document}